\documentclass{article}

\usepackage{microtype}
\usepackage{graphicx}
\usepackage{subfigure}
\usepackage{booktabs} 

\usepackage{hyperref}

\usepackage[accepted]{icml2024}

\usepackage{amsmath}
\usepackage{amssymb}
\usepackage{mathtools}
\usepackage{amsthm}

\usepackage{colortbl}
\usepackage{xcolor}
\usepackage{multirow}
\usepackage{makecell}
\usepackage{bm}
\usepackage{booktabs}
\usepackage{color}
\usepackage{graphicx}

\usepackage[capitalize,noabbrev]{cleveref}

\theoremstyle{plain}

\theoremstyle{definition}

\theoremstyle{remark}

\usepackage[textsize=tiny]{todonotes}

\icmltitlerunning{iVPT: Improving Task-relevant Information Sharing in Visual Prompt Tuning by Cross-layer Dynamic Connection}

\begin{document}

\twocolumn[

\icmltitle{iVPT: Improving Task-relevant Information Sharing in \\ Visual Prompt Tuning by Cross-layer Dynamic Connection}



\icmlsetsymbol{equal}{*}

\begin{icmlauthorlist}
\icmlauthor{Nan Zhou}{}
\icmlauthor{Jiaxin Chen}{}
\icmlauthor{Di Huang}{}

\end{icmlauthorlist}


\icmlcorrespondingauthor{Di Huang}{dhuang@buaa.edu.cn}

\icmlkeywords{Machine Learning, ICML}

\vskip 0.3in
]



\printAffiliationsAndNotice{}  

\begin{abstract}
Recent progress has shown great potential of visual prompt tuning (VPT) when adapting pre-trained vision transformers to various downstream tasks. However, most existing solutions independently optimize prompts at each layer, thereby neglecting the usage of task-relevant information encoded in prompt tokens across layers. Additionally, existing prompt structures are prone to interference from task-irrelevant noise in input images, which can do harm to the sharing of task-relevant information. In this paper, we propose a novel VPT approach, \textbf{iVPT}. It innovatively incorporates a cross-layer dynamic connection (CDC) for input prompt tokens from adjacent layers, enabling effective sharing of task-relevant information. Furthermore, we design a dynamic aggregation (DA) module that facilitates selective sharing of information between layers. The combination of CDC and DA enhances the flexibility of the attention process within the VPT framework. Building upon these foundations, iVPT introduces an attentive reinforcement (AR) mechanism, by automatically identifying salient image tokens, which are further enhanced by prompt tokens in an additive manner. Extensive experiments on 24 image classification and semantic segmentation benchmarks clearly demonstrate the advantage of the proposed iVPT, compared to the state-of-the-art counterparts.

\end{abstract}

\section{Introduction}
The continuous emergence of pre-trained foundation models has greatly advanced the fields of natural language processing (NLP) \cite{bert, gpt3} and computer vision \cite{moco, sam}, and it is becoming an urgent demand to adapt these models and unleash their potential, aiming at facilitating more downstream applications. Conventional approaches generally focus on full fine-tuning \cite{xuhong2018L2SP, drtune}, however they suffer from rapidly growing model sizes \cite{vit, swin, vmoe}. Recently, many efforts have been made to explore efficient alternatives dubbed as parameter efficient fine-tuning (PEFT) \cite{lora, adaptformer}. As one of the most prominent representatives in this domain, prompt tuning, which originates in NLP, only fine-tunes a small portion of prompt parameters while freezing the majority of the pre-trained model, thus significantly decreasing the learning and storage cost \cite{prompt_survey}. Inspired by its success in NLP, VPT \cite{jia2022vpt} makes the first investigation on prompt tuning in the computer vision community and adapts pre-trained transformers to major vision tasks, reporting promising results.

\begin{figure}[t]
  \centering 
  \includegraphics[width=80mm]{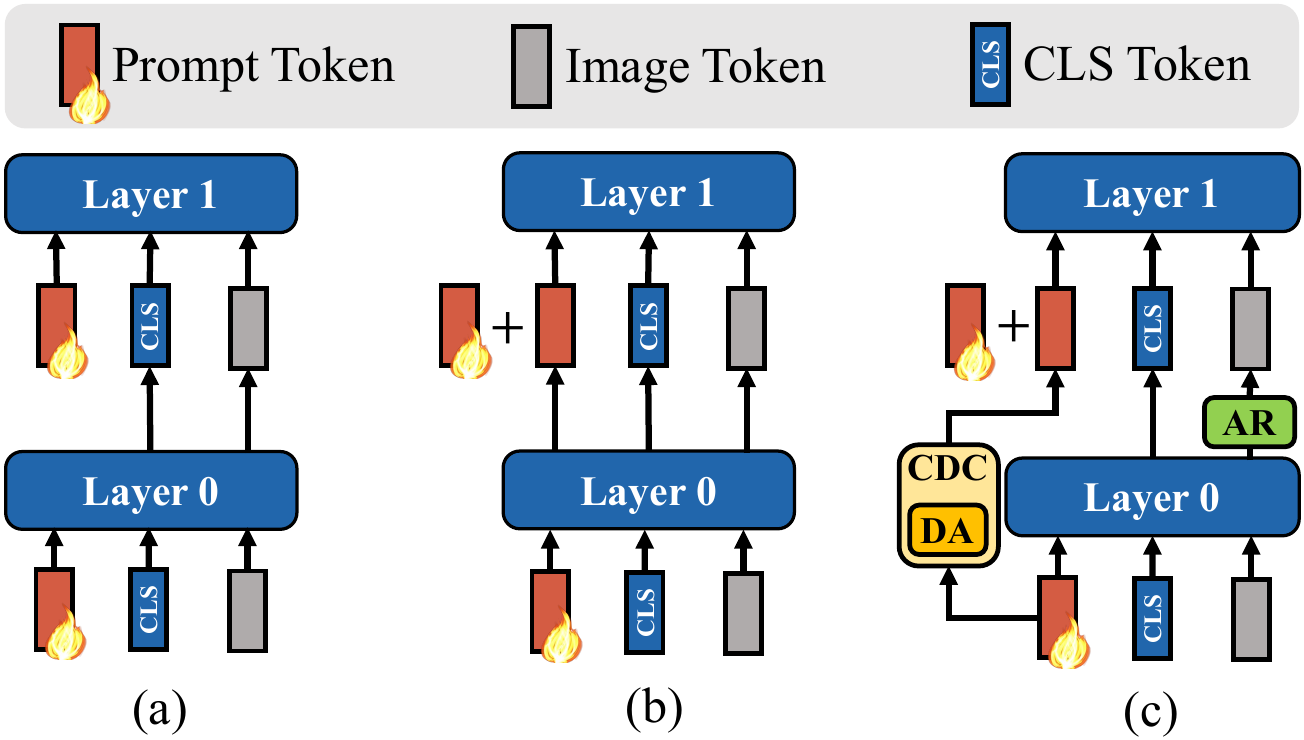}
  \vskip -0.1in

  \caption{Comparison of distinct VPT approaches. (a) VPT-deep truncates output prompt tokens and independently learns prompt tokens across layers. (b) Existing attempts preserve output prompt tokens and learn prompts on them. In contrast, (c) iVPT builds cross-layer dynamic connection (CDC) with dynamic aggregation (DA) on input prompt tokens, benefiting task-relevant information sharing, and introduces the attentive reinforcement (AR) module to highlight salient image regions.} 
  \label{compare_arch_simple}
\end{figure}

The common practice of VPT is to insert learnable prompt tokens into certain transformer layers, \emph{a.k.a.} VPT-shallow or VPT-deep, and these prompt tokens interact with image tokens, accomplishing model adaptation without modifying the pre-trained model. The follow-up works further improve this pipeline from three main aspects: (1) reducing learnable parameters in prompt tuning \cite{e2vpt}, (2) developing input-dependent prompt mechanisms to deal with large variations of images \cite{dam-vp, l2p}, and (3) designing various prompt structures to explore prompt correlations \cite{progressive, express, gate}. Unfortunately, most of the approaches \cite{jia2022vpt, e2vpt} independently optimize prompt tokens by truncating output prompt tokens at each layer, as shown in Fig.~\ref{compare_arch_simple} (a), and it substantially restrains sharing task-relevant information in prompts across different layers, which is considered to contribute to adaptation performance. A few attempts \cite{progressive, express, gate} preserve output prompt tokens at each layer (see Fig.~\ref{compare_arch_simple} (b)) and indeed alleviate the issue of inter-layer information sharing to some extent, but this strategy makes them susceptible to interference from input images as the output prompts involve cross-attention with image tokens.

To address the limitations above, we propose a novel PEFT approach, namely iVPT, which improves the VPT framework by inter-layer task-relevant information sharing in adapting transformer-based pre-trained foundation vision models. As displayed in Fig.~\ref{compare_arch_simple} (c), we firstly introduce a new prompt tuning architecture dubbed as cross-layer dynamic connection (CDC) to generate input prompt tokens in a certain layer based on those from its previous layers. Since prompts primarily encode task-relevant clues as evaluated in \cite{prompt_transferability, multitask_prompt}, CDC benefits sharing such information between prompt tokens across layers while avoiding the disturbance of noise from input images, due to the data-independent spirit of the connection. Moreover, considering that not all task-relevant information is transferable across layers, we design a dynamic aggregation (DA) module. It selectively fulfills transfer through a set of learnable parameters, enabling nuanced control of information to be shared. Based on CDC, we further present an innovative prompting mechanism named Attentive Reinforcement (AR) to optimize the process of encoding prompts into image tokens, which automatically identifies and enhances salient image regions according to attention weights with learnable prompt tokens. AR bypasses the need for additional query processes \cite{l2p} or external data assistance \cite{dam-vp}, adaptively guiding the pre-trained model to impose higher attentions on task-relevant foreground regions than those irrelevant ones.

The main contributions are summarized as follows: 

1) We propose a novel VPT approach, namely iVPT, which benefits task-relevant information sharing between inter-layer prompt tokens by a specially designed prompt structure (\emph{i.e.} CDC) and aggregation module (\emph{i.e.} DA). Theoretical and empirical analyses are both provided, highlighting its superiority compared to other prompt structures.

2) We design an innovative prompt mechanism (\emph{i.e.} AR) to identify and reinforce salient image regions, guiding the pre-trained model to concentrate more on task-relevant foreground areas.

3) We conduct extensive evaluation on 24 vision tasks for image classification and semantic segmentation, and demonstrate that iVPT achieves new state-of-the-art performance, exceeding that of the prompt-based counterparts and other representative PEFT approaches.

\section{Related Work}
\subsection{Parameter Efficient Fine-Tuning}
\label{PEFT}
The explosive growth of the number of pre-trained model parameters incurs heavy computation and storage overhead during the model 
adaptation process. Various PEFT strategies have been proposed for
a trade-off between the adaptation performance and cost. 
The partial fine-tuning based approaches \cite{bitfit,sct} train a 
subset of the model parameters and freeze the rest. 
Other approaches design lightweight adaptation modules, which consist bottleneck-like architectures, 
\emph{a.k.a.} adapters \cite{adaptformer,adapter_nlp,bi}, low-rank adaptation module \cite{lora}, prompts \cite{jia2022vpt}, scale\&shift operations \cite{ssf}, or normalizing flow layers \cite{snf}. 
Such modules are inserted into the frozen pre-trained model and optimized by downstream data.

\begin{figure*}
    \centering 
    \includegraphics[width=145mm]{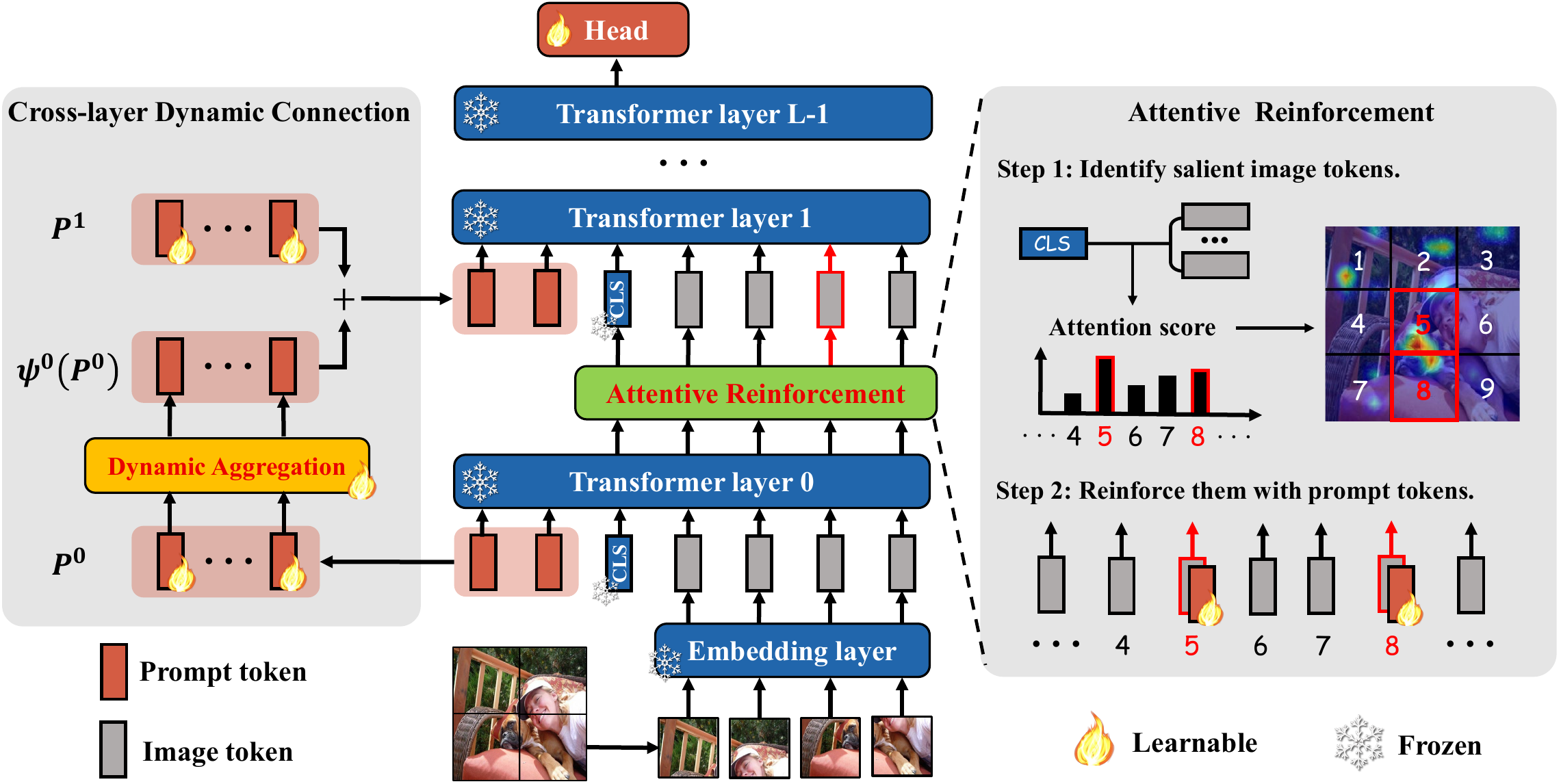}
    \caption{Illustration on the proposed iVPT approach. 
    1) Cross-layer dynamic connection (CDC): dynamically aggregates (DA) and connects the input prompt tokens at adjacent layers. 
    2) Attentive reinforcement (AR): first identifies salient image tokens according to the attention weights with cls token and then enhances salient image tokens with learnable prompt tokens.}
    \label{pipeline}
  \end{figure*}

\subsection{Visual Prompt Tuning}
\label{prompt_related}
The concept of prompt was originally born in the field of NLP \cite{prompt_survey}. By designing text prompts, downstream tasks 
are reformulated to look more like those in the pre-training phase, which can be solved by pre-trained models without modifying their parameters \cite{hard_prompt_2}. 
VPT \cite{jia2022vpt} and VP \cite{vp} are two pioneers that try prompt tuning on general vision tasks and learn visual prompts at token-level and pixel-level respectively. E$^2$VPT \cite{e2vpt} introduces prompts to the key and value matrix and designs a prompt pruning mechanism. 
DAM-VP \cite{dam-vp} divides downstream dataset with the help of clustering algorithms and individually learns prompts for each sub-dataset. Pro-tuning \cite{nie2022pro_tuning} and L2P \cite{l2p} introduces prompt generator network and query process to select key-prompt pairs, respectively. 
However, most methods \cite{jia2022vpt,e2vpt} learn prompt tokens at different layers in an isolated way, ignoring to explore the shareable task-relevant information between layers, thus leaving much room for improvements. 
Other prompt mechanisms require specific network design \cite{nie2022pro_tuning, pgn}, extra query processes \cite{l2p,dualprompt, dam-vp} or external data assistance \cite{dam-vp}, bringing storage and computational overhead. 

\subsection{Prompt Structures in Visual Prompt Tuning}
Some works design various prompt structures for visual prompt tuning. 
ProVP \cite{progressive} designs a progressive prompt structure on output prompt tokens to strengthen the interaction between prompts of different layers. 
EXPRESS \cite{express} inherits the output prompt tokens and inserts additional prompt tokens before the layer normalization, QKV projection and self-attention layers. 
GaPT \cite{gate} introduces a gate mechanism to automatically find the optimal insertion position of prompt tokens. 
These methods benefit the sharing of task-relevant information across layers, but are susceptible to task-irrelevant information in the input image. 
Compared with existing prompt-based methods, iVPT enables task information sharing without interference from image noise via CDC and DA, while AR highlights the salient regions without data assistance or additional query process. 

\section{The Proposed Method}
\subsection{Preliminary}
\textbf{Vision transformer (ViT).} Given an image $\bm{I} \in \mathbb{R}^{H \times W \times 3}$, ViT \cite{vit} firstly splits it into $M$ non-overlapping 
image patches $\{\bm{I}_{i}\}_{i=1}^{M}$, which are successively projected to $D$-dimensional image tokens $\{\bm{z}_i^{0} \in \mathbb{R}^{D}\}_{i=1}^{M}$. By rearranging $\{\bm{z}_i^{0}\}_{i=1}^{M}$ into $\bm{Z}^0=[\bm{z}_1^0, \cdots, \bm{z}_M^0] \in \mathbb{R}^{M \times D}$, adding the positional embedding and concatenating a class token $\bm{z}_{\rm cls}^0 \in \mathbb{R}^{D}$, the input is formulated as $[\bm{z}_{\rm cls}^0, \bm{Z}^0] \in \mathbb{R}^{(1+M) \times D}$, which is fed to $L$ transformer layers $\{\mathcal{T}^l\}_{l=0}^{L-1}$ as follows:
\begin{equation}
\left[\bm{z}_{\rm cls}^{l+1}, \ \bm{Z}^{l+1}\right] = \mathcal{T}^l\left(\left[\bm{z}_{\rm cls}^{l}, \ \bm{Z}^{l}\right]\right); l=0,\cdots,L-1.
\label{ori_inference}
\end{equation}

\textbf{Visual prompt tuning (VPT).} VPT \cite{jia2022vpt} is the pioneer that aims to efficiently adapt pre-trained ViTs to downstream tasks. It basically introduces a small amount of prompt tokens, and subsequently optimizes them whilst keeping the pre-trained model frozen. VPT mainly has two variants, one of which denoted as VPT-shallow only inserts prompt tokens to the first transformer layer, and the other one dubbed VPT-deep instead inserts prompt tokens to all layers. By employing learnable prompt tokens to interact with image tokens in more layers, VPT-deep usually achieves better performance. Formally, we introduce $N$ prompt tokens at layer $l$ denoted as $\bm{P}^{l}=[\bm{p}_1^{l}, \cdots, \bm{p}_N^{l}] \in \mathbb{R}^{N \times D}$, based on which the output in layer $l$ as shown in Eq.~\ref{ori_inference} is rewritten as below: 
\begin{equation}
\begin{aligned}
  \left[\bm{z}_{\rm cls}^{l+1}, \ \_\_, \ \bm{Z}^{l+1}\right] = \mathcal{T}^l\left(\left[\bm{z}_{\rm cls}^{l}, \ \bm{P}^{l}, \ \bm{Z}^{l}\right]\right).
  \label{vpt_inference}
\end{aligned}
\end{equation}
Note that VPT-deep does not preserve the output corresponding to the prompt token $\bm{P}^{l}$ thus being omitted denoted by the symbol `$\_\_$' as in Eq.~\eqref{vpt_inference}. Alternatively, VPT-deep employs new learnable prompt tokens $\bm{P}^{l+1}$ for computation in subsequent layer $l+1$, thus being isolated to $\bm{P}^{l}$.

\subsection{Approach Overview}
We propose a novel visual prompt tuning approach for adapting pre-trained vision transformers, 
namely iVPT. As shown in Fig.~\ref{pipeline}, given an input image, the embedding layer 
converts it to a sequence of image tokens and then it concatenates the class token and prompt tokens to form the input sequence. 
In contrast to VPT, we add cross-layer dynamic connection (CDC) with dynamic aggregation (DA) on the input prompt tokens at adjacent layers. Such a structure facilitates the task-relevant information encoded in prompt token sharing across layers and leads to a more flexible attention process. 
Attentive reinforcement (AD) further utilizes the benefits of CDC and DA and reinforces the image tokens according to the flexible attention weights.

\subsection{Cross-layer Dynamic Connection}

Inspired by the success of residual learning in deep convolution neural networks \cite{he2016resnet}, we introduce a cross-layer dynamic connection (CDC) between the input prompt tokens from adjacent layers, formulated as below:
\begin{align}
  \left[\bm{z}_{\rm cls}^{l+1}, \ \_\_, \ \bm{Z}^{l+1}\right] &=  \mathcal{T}^l\left(\left[\bm{z}_{\rm cls}^{l}, \ \hat{\bm{P}}^{l}, \ \bm{Z}^{l}\right]\right),
  \label{no_dyn_residual_}
\end{align}
where 
\begin{align}
  \hat{\bm{P}}^{l} = \begin{cases}
    \bm{P}^{0}, & l=0 \\
    \bm{P}^{l} + \bm{P}^{l-1}, & l=1,\cdots,L-1\\
  \end{cases}
  \label{no_dyn_residual} 
\end{align}
As shown in Eq.~\eqref{no_dyn_residual_} and ~\eqref{no_dyn_residual}, the input prompt $\hat{\bm{P}}^{l}$ in layer $l$ is derived from the prompt $\bm{P}^{l}$ by adding an additional prompt $\bm{P}^{l-1}$ from previous layer $l-1$. In contrast with the isolated optimization commonly seen in VPT, as CDC allows $\bm{P}^{l}$ to learn from $\bm{P}^{l-1}$, facilitating the transfer of task-relevant information encoded by $\bm{P}^{l-1}$ to $\bm{P}^{l}$.

However, the one-to-one additive relationship between adjacent layers in Eq.~\eqref{no_dyn_residual} imposes strong constraints. On the one hand, layers have different preferences for task-relevant information, necessitating selective determination of the extent and intensity of information sharing.
On the other hand, the $i$-th prompt token $\bm{p}_i^{l-1}$ in layer $l-1$ sharing information only with its counterpart $\bm{p}_i^{l}$ from subsequent layer introduces position biases. To address these issues, we introduce a dynamic aggregation (DA) module, which dynamically aggregates all prompt tokens in layer $l-1$ for each prompt token in layer $l$ via a learnable linear weighting function, based on which Eq.~\eqref{no_dyn_residual} is reformulated as follows: 
\begin{equation}
\begin{array}{rl}
  \hat{\bm{P}}^{l} = \begin{cases}
    \bm{P}^{0}, & l=0 \\
    \bm{P}^{l} + \psi^{l-1}(\bm{P}^{l-1}), & l=1,\cdots,L-1\\
  \end{cases} \\
\end{array}
\label{dyn_residual}
\end{equation}
where 
\begin{align}
  \psi^{l-1}(\bm{P}^{l-1}) &= \left[\sum_{j=1}^{N} \gamma_{i,j}^{l-1} * \bm{p}^{l-1}_j\right]_{i=1}^N.
\end{align}
The $N\times N$ learnable weights $\{\gamma_{i,j}^{l-1}\}_{i,j=1}^N$ in DA eliminate the positional constraints, allowing each prompt token to selectively aggregate task-relevant information from $N$ prompt tokens in the previous layer.

\subsection{Understanding Cross-layer Dynamic Connection}
\label{proof_residual} 
In this section, we aim to clarify the functionality of the CDC and the role of DA in relation to attention mechanisms.

For clarity, we temporarily disregard the notation for layer normalization and concentrate on the interaction between the class token and the prompt tokens at layer $l$. Drawing on the analysis in \cite{express}, we conceptualize the multi-head self-attention (MSA) with $N_h$ heads as a weighted summation over all value tokens. In traditional VPT, the attention weight $w_i$ between the class token and $i$-th prompt token is determined as follows:
\begin{align}
  w_i \propto {\rm exp} \left(\frac{\bm{q}^T \bm{k}_i}{\sqrt{D/N_h}}\right).
  \label{ori_atten_weight}
\end{align}
Expending on $\bm{k}_i$ as illustrated in Eq.~\eqref{ori_atten_weight}, and in scenarios where DA is not factored (\emph{i.e.} $\psi$ is an identity mapping), the attention weight in CDC is redefined as:
\begin{align}
  \tilde{w}_i &\propto {\rm exp} \left(\frac{\bm{q}^T \bm{W}_K^l (\bm{p}^l_i+\bm{p}^{l-1}_i)}{\sqrt{D/N_h}}\right) \label{ours_attn} \\
  &=\mathrm{exp} \left(\frac{\bm{q}^T \bm{W}_K^l \bm{p}^l_i}{\sqrt{D/N_h}}\right) \mathrm{exp} \left(\frac{\bm{q}^T \bm{W}_K^l \bm{p}^{l-1}_i}{\sqrt{D/N_h}}\right) \\
  &\propto w_{i}*\tilde{\alpha}_{i}, \label{ours_attn_final}
\end{align}
where $\bm{W}_K^l$ denotes the key projection matrix at layer $l$. 
The input prompt token $\bm{p}^{l-1}_i$ from the previous layer, introduced by CDC, acts as the re-weighting term $\hat{\alpha}_{i}$, aiding in the adjustment of attention weights.
Furthermore, with the incorporation of DA, the term $\bm{p}^{l-1}_i$ in Eq.~\eqref{ours_attn} is substituted by $\sum_{j=1}^{N} \gamma_{i,j}^{l-1} \bm{p}^{l-1}_j$, allowing Eq.~\eqref{ours_attn} to be revised as:
\begin{align}
  \tilde{w}_i &\propto {\rm exp} \left(\frac{\bm{q}^T \bm{W}_K^l \left(\bm{p}^l_i+ \sum_{j=1}^{N} \gamma_{i,j}^{l-1} \bm{p}^{l-1}_j\right)}{\sqrt{D/N_h}}\right) \label{ours_dyn_benefit}\\
  &=\mathrm{exp} \left(\frac{\bm{q}^T \bm{W}_K^l \bm{p}^l_i}{\sqrt{D/N_h}}\right) \prod_{j=1}^{N} \mathrm{exp} \left(\frac{\bm{q}^T \bm{W}_K^l \gamma_{i,j}^{l-1} \bm{p}^{l-1}_j}{\sqrt{D/N_h}}\right) \\
  &\propto w_{i}* \prod_{j=1}^{N} \tilde{\alpha}_{j}.  \label{ours_attn_dyn}
\end{align}
By comparing Eq.~\eqref{ours_attn_dyn} with Eq.~\eqref{ours_attn_final}, it becomes evident that the re-weighting term is no longer solely reliant on the prompt token at the corresponding position in the previous layer. Instead, DA allow all $N$ prompt tokens from the previous layer selectively contribute to the re-weighting process. The proposed DA further relaxes the constraints on prompt tokens, resulting in a more adaptable and flexible attention mechanism.

\begin{table*} \scriptsize 
  \centering 
  \setlength{\tabcolsep}{2.5pt}
  \caption{Comparison of the top-1 accuracies (\%) and learnable parameters (M) by various methods on VTAB-1k \cite{vtab} with ViT-B/16 supervised pre-trained on 
  ImageNet-21k. The best and second best results are highlighted in \textbf{bold} and \underline{underline}. `-' indicates the result was not reported and `$^\dagger$' indicates results reported in \cite{sct}.  } 
  \vskip 0.1in
  \begin{tabular}{l c c c c c c c c | c c c c c | c c c c c c c c c | c c} 
      \toprule 
      ~ & \multicolumn{8}{c}{\textbf{Natural}} & \multicolumn{5}{c}{\textbf{Specialized}} & \multicolumn{9}{c}{\textbf{Structured}} \\
      \cmidrule(lr){2-9}\cmidrule(lr){10-14}\cmidrule(lr){15-23}
      \makecell[l]{Method} & \rotatebox{90}{CIFAR100} & \rotatebox{90}{Caltech101} & \rotatebox{90}{DTD} & \rotatebox{90}{Flowers102} & 
      \rotatebox{90}{Pets} & \rotatebox{90}{SVHN} & \rotatebox{90}{Sun397} & \rotatebox{90}{Mean Acc.} & 
      \rotatebox{90}{Patch Came.} & \rotatebox{90}{EuroSAT} & \rotatebox{90}{Resisc45} & \rotatebox{90}{Retinopathy} & \rotatebox{90}{Mean Acc.} & 
      \rotatebox{90}{Clevr/count} & \rotatebox{90}{Clevr/dist.} & \rotatebox{90}{DMLab} & \rotatebox{90}{KITTI/dist.} & 
      \rotatebox{90}{dSprites/loc} & \rotatebox{90}{dSprites/ori} & \rotatebox{90}{NORB/azi} & \rotatebox{90}{NORB/ele} & \rotatebox{90}{Mean Acc.} & 
      \rotatebox{90}{Group Mean Acc.} & \rotatebox{90}{Mean params.}\\
      \midrule
      \makecell[l]{Full fine-tuning} & 68.9 & 87.7 & 64.3 & 97.2 & 86.9 & 87.4 & 38.8 & 75.9 & 79.7 & 95.7 & 84.2 & 73.9 & 83.4 & 56.3 & 58.6 & 41.7 & 65.5 & 57.5 & 46.7 & 25.7 & 29.1 & 47.6 & 68.90 & 85.84\\
      \makecell[l]{Linear probing} & 63.4 & 85.0 & 63.2 & 97.0 & 86.3 & 36.6 & 51.0 & 68.9 & 78.5 & 87.5 & 68.6 & 74.0 & 77.2 & 34.3 & 30.6 & 33.2 & 55.4 & 12.5 & 20.0 & 9.6 & 19.2 & 26.9 & 57.67 & \textbf{0.04}\\
      \makecell[l]{Adapter} & 74.1 & 86.1 & 63.2 & 97.7 & 87.0 & 34.6 & 50.8 & 70.5 & 76.3 & 88.0 & 73.1  & 70.5 & 77.0 & 45.7 & 37.4 & 31.2 & 53.2 & 30.3 & 25.4 & 13.8 & 22.1 & 32.4 & 59.97 & 0.27\\
      \makecell[l]{Bias} & 72.8 & 87.0 & 59.2 & 97.5 & 85.3 & 59.9 & 51.4 & 73.3 & 78.7 & 91.6 & 72.9 & 69.8 & 78.3 & 61.5 & 55.6 & 32.4 & 55.9 & 66.6 & 40.0 & 15.7 & 25.1 & 44.1 & 65.23 & 0.14\\
      \makecell[l]{LoRA$^\dagger$}  & 67.1 & 91.4 & 69.4 & 98.8 & 90.4 & 85.3 & 54.0 & 79.5 & 84.9 & 95.3 & 84.4 & 73.6 & 84.6 & 82.9 & \textbf{69.2} & 49.8 & 78.5 & 75.7 & 47.1 & 31.0 & 44.0 & 59.8 & 74.63 & 0.43 \\
      \makecell[l]{AdaptFormer$^\dagger$} & 73.5 & 91.5 & 70.8 & 98.9 & 91.1 & 87.8 & 54.3 & 81.1 & 82.3 & 94.9 & 86.9 & \underline{76.3} & 85.1 & 82.3 & 66.3 & 51.0 & 78.8 & 76.5 & 46.2 & 32.3 & 40.1 & 59.2 & 75.13 & 0.22\\
      \makecell[l]{SSF} & 69.0 & 92.6 & \textbf{75.1} & \textbf{99.4} & \textbf{91.8} & 90.2 & 52.9 & 81.6 & 87.4 & 95.9 & \textbf{87.4} & 75.5 & \underline{86.6} & 75.9 & 62.3 & \textbf{53.3} & 80.6 & 77.3 & \underline{54.9} & 29.5 & 37.9 & 59.0 & 75.73 & 0.24\\

      \makecell[l]{SCT} & 75.3 & 91.6 & 72.2 & 99.2 & 91.1 & \textbf{91.2} & 55.0 & 82.2 & 85.0 & 96.1 & 86.3 & 76.2 & 85.9 & 81.5 & 65.1 & 51.7 & 80.2 & 75.4 & 46.2 & \underline{33.2} & \underline{45.7} & 59.9 & 76.00 & 0.15 \\ 
      \makecell[l]{FacT} & 70.6 & 90.6 & 70.8 & 99.1 & 90.7 & 88.6 & 54.1 & 80.6 & 84.8 & 96.2 & 84.5 & 75.7 & 85.3 & 82.6 & \underline{68.2} & 49.8 & 80.7 & 80.8 & 47.4 & \underline{33.2} & 43.0 & 60.7 & 75.53 & \underline{0.11} \\
      \makecell[l]{SNF} & \textbf{84.0} & \underline{94.0} & \underline{72.7} & \underline{99.3} & 91.3 & \underline{90.3} & 54.9 & \textbf{83.8} & 87.2 & \textbf{97.3} & 85.5 & 74.5 & 86.1 & 82.3 & 63.8 & 49.8 & \textbf{82.5} & 75.8 & 49.2 & 31.4 & 42.1 & 59.6 & 76.50 & 0.29 \\
      \makecell[l]{Bi-Adapter} & 74.1 & 92.4 & 72.1 & \underline{99.3} & \underline{91.6} & 89.0 & \underline{56.3} & 82.1 & \textbf{88.2} & 95.2 & 86.0 & 76.2 & 86.4 & \underline{83.9} & 63.6 & \underline{53.0} & 81.4 & \textbf{86.2} & 54.8 & \textbf{35.2} & 41.3 & \textbf{62.4} & \underline{76.97} & 0.64 \\
      \makecell[l]{SPT} & 73.5 & 93.3 & 72.5 & \underline{99.3} & 91.5 & 87.9 & 55.5 & 81.9 & 85.7 & 96.2 & 85.9 & 75.9 & 85.9 & \textbf{84.4} & 67.6 & 52.5 & \underline{82.0} & 81.0 & 51.1 & 30.2 & 41.3 & \underline{61.3} & 76.37 & 0.54\\

      \midrule
      \multicolumn{2}{c}{\textcolor{gray}{Prompt-based methods:}}\\
      \makecell[l]{VPT} & 78.8 & 90.8 & 65.8 & 98.0 & 88.3 & 78.1 & 49.6 & 78.5 & 81.8 & 96.1 & 83.4 & 68.4 & 82.4 & 68.5 & 60.0 & 46.5 & 72.8 & 73.6 & 47.9 & 32.9 & 37.8 & 55.0 & 71.97 & 0.60\\
      \makecell[l]{EXPRESS} & 78.0 & 89.6 & 68.8 & 98.7 & 88.9 & 81.9 & 51.9 & 79.7 & 84.8 & 96.2 & 80.9 & 74.2 & 84.0 & 66.5 & 60.4 & 46.5 & 77.6 & 78.0 & 49.5 & 26.1 & 35.3 & 55.0 & 72.90 & 0.98\\
      \makecell[l]{DAM-VP}   & - & - & - & - & - & - & - & 81.3 & - & - & - & - & 83.8 & - & - & - & - & - & - & - & - & 54.3 & 73.13 & 2.52\\
      \makecell[l]{E$^2$VPT} & 78.6 & 89.4 & 67.8 & 98.2 & 88.5 & 85.3 & 52.3 & 80.0 & 82.5 & \underline{96.8} & 84.8 & 73.6 & 84.4 & 71.7 & 61.2 & 47.9 & 75.8 & 80.8 & 48.1 & 31.7 & 41.9 & 57.4 & 73.93 & 0.27\\	
      \rowcolor{gray!20} \makecell[l]{iVPT~~~~~~~~~~~~~~~~~~} & \underline{82.7} & \textbf{94.2} & 72.0 & 99.1 & \textbf{91.8} & 88.1 & \textbf{56.6} & \underline{83.5} & \underline{87.7} & 96.1 & \underline{87.1} & \textbf{77.6} & \textbf{87.1} & 77.1 & 62.6 & 49.4 & 80.6 & \underline{82.1} & \textbf{55.3} & 31.8 & \textbf{47.6} & 60.8 & \textbf{77.13} & 0.60\\	
      \bottomrule 
    \end{tabular}
  
  \label{vtab_sup_compare}
\end{table*}

\subsection{Attentive Reinforcement}

From Eq.~\eqref{ours_attn_dyn}, we conjecture the flexible attention process implemented by CDC and DA can help the pre-trained model to focus on salient regions that are helpful for downstream tasks.
We propose the attentive reinforcement (AR) module, which mainly consists of two steps. 
Firstly, for a total of $M$ output image tokens $\bm{Z}^{l+1} \in \mathbb{R}^{M \times D}$ in layer $l$, we select the most salient $k$ image tokens according to their attention weights $\tilde{\bm w}^l \in \mathbb{R}^M$ with the class token. 
Secondly, we additionally introduce $k$ learnable prompt tokens $\bm{P}^{l}_{\rm re} \in \mathbb{R}^{{k} \times D}$ 
and add it to the most salient image tokens formulated as below: 
\begin{align}
\Omega^l &= {\rm Topk} (\tilde{\bm w}^l, {k}), \\
\tilde{\bm{Z}}^{l+1}\left[\Omega^l\right] &= \bm{Z}^{l+1}\left[\Omega^l\right] + \bm{P}^{l}_{\rm re},
\end{align}
where ${\rm Topk}\left(\tilde{\bm w}^l, \* {k}\right)$ returns the descending order indices of the largest $k$ elements in $\tilde{\bm w}^l$ and 
$[\, \cdot \,]$ represents the indexing operation. 
The reinforced image tokens $\tilde{\bm{Z}}^{l+1}$ together with the class token $\bm{z}_{\rm cls}^{l+1}$ and the prompt tokens $\hat{\bm{P}}^{l+1}$ in CDC are subsequently input into the next transformer layer $l+1$.

\section{Experimental Results and Analysis} 
\subsection{Experimental Settings}
\textbf{Datasets.}
We evaluate the proposed iVPT method on the widely used VTAB-1k \cite{vtab}, Fine-grained visual classification (FGVC) and ADE-20k \cite{ade} benchmark. 
VTAB-1k is a collection of 19 public datasets for diverse 
vision tasks, which are organized into three groups: \textbf{1)} \emph{Natural} tasks that classify natural images captured
by standard cameras; \textbf{2)} \emph{Specialized} tasks that process images captured via specialized equipment, such as the medical and satellite images; 
\textbf{3)} \emph{Structured} tasks that require geometric comprehension such as object counting. FGVC contains five fine-grained image datasets including CUB-200-2011 \cite{cub}, NABirds \cite{nabird}, Oxford Flowers \cite{flowers}, Stanford Dogs \cite{dog} and Stanford Cars \cite{cars}.

\textbf{Implementation Details.}
By following \cite{jia2022vpt}, we adopt ViT-B/16 \cite{vit} supervised pre-trained on ImageNet-21k \cite{deng2009imagenet} as the backbone in 
most experiments. To comprehensively evaluate our method, we also provide experimental results by our method with different backbones and distinct pre-training strategies in Sec ~\ref{Generalizability}. 
During training, we use a single NVIDIA-2080Ti GPU and conduct a linear warm-up training for the first 10 epochs and adopt the SGD optimizer \cite{bottou2010SGD} with a cosine decay scheduler for the remaining 90 epochs.
Similar to \cite{jia2022vpt,express,dam-vp}, we utilize the random cropping and horizontal flip for data augmentation with an image size of
224$\times$224 during training, and apply the center cropping during testing. 

Please refer to the \emph{appendix} for more details about hyper-parameters and datasets.

\begin{table}[t]\small  
	\centering 
 \caption{Ablation study on the main components. CDC: cross-layer dynamic connection. DA: dynamic aggregation. AR: attentive reinforcement.}
 \vskip 0.1in 
 \setlength{\tabcolsep}{4.5pt}
  \begin{tabular}{c c c c c c c} 	
	\toprule
    CDC & DA & AR & Natural & Specialized & Structured & Mean \\
    \midrule
    \multicolumn{3}{c}{\textcolor{gray}{VPT baseline $\rightarrow$}} & 78.48 & 82.43 & 54.98 & 71.97\\
    \checkmark & & & 82.57 & 85.30 & 58.37 & 75.41\\
    \checkmark & \checkmark &  & 82.86 & 85.99 & 59.19 & 76.01\\
    \checkmark & \checkmark & \checkmark & \textbf{83.50} & \textbf{87.13} & \textbf{60.81} & \textbf{77.13}\\
    \bottomrule 
\end{tabular}
    \label{diff_components}
  \end{table}

  \begin{figure*}[t]
    \centering 
    \includegraphics[width=168mm]{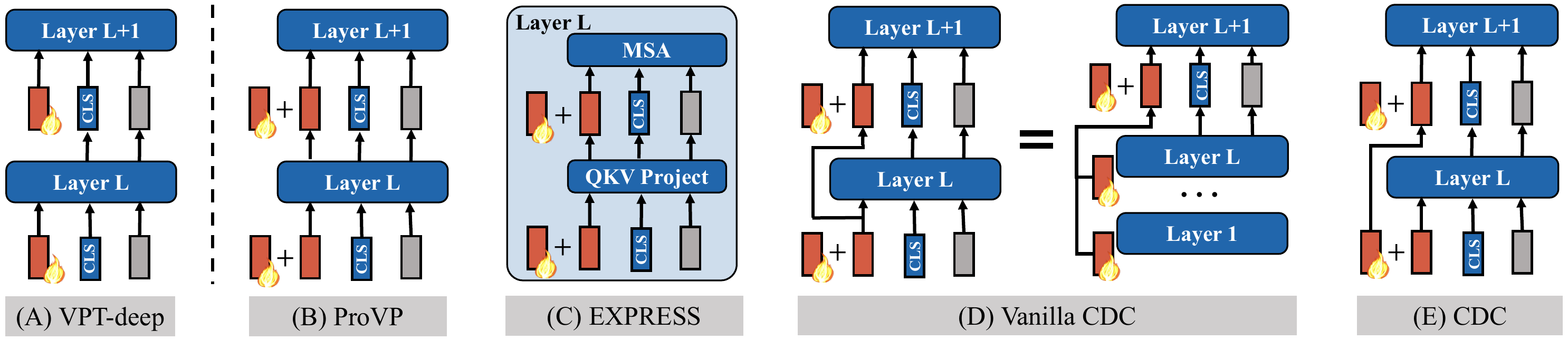}
    \vskip -0.1in
    \caption{Illustration on different prompt structures. (A) VPT-deep \cite{jia2022vpt}. (B) ProVP \cite{progressive}, which preserves the output prompt tokens and adds learnable prompt tokens. (C) EXPRESS \cite{express}, which preserves the output prompt tokens and adds prompt tokens before LN, QKV projection and MSA layers. (D) Vanilla CDC, which transfers the input prompt token from all preceding layers to the current layer. (E) CDC, which transfers the input prompt token solely from previous layer to the current layer.} 
    \label{arch_ill}
  \end{figure*}

\subsection{Comparison with the State-of-the-art}
We compare our approach with the following state-of-the-arts methods. \textbf{1)} The baseline methods including Full fine-tuning and Linear probing. 
\textbf{2)} The partial fine-tuning methods Bias \cite{bitfit} and SCT \cite{sct} that only updates the bias terms and the important channels. \textbf{3)} The adapter-based methods including Adapter \cite{adapter_nlp}, AdaptFormer \cite{adaptformer}, Bi-Adapter \cite{bi}, SSF \cite{ssf} and SNF \cite{snf}
which insert lightweight adapters, scaling/shifting operators and normalizing flow layers into the frozen pre-trained model. \textbf{4)} The decomposition method LoRA \cite{lora}, SPT \cite{spt} and FacT \cite{facT}. \textbf{5)} The prompt-based methods including VPT \cite{jia2022vpt}, EXPRESS \cite{express}, DAM-VP \cite{dam-vp} and E$^2$VPT \cite{e2vpt}. 
  
As summarised in Table~\ref{vtab_sup_compare}, the baseline method Linear probing uses the least number of learnable parameters but performs the worst, showing that only optimizing the task head is insufficient to achieve a competitive performance. 
Full fine-tuning substantially promotes the accuracy, but incurs a huge parameter overhead, indicating the necessity of designing more efficient model adaptation methods. By introducing the CDC, DA and the AR module, iVPT significantly boosts the accuracy of Full fine-tuning as well as the compared prompt-based methods by adopting negligible learnable parameters. We also compare to other representative PEFT methods, and our method reaches the highest accuracy with comparable amount of learnable parameters.

\subsection{Ablation Study and Analysis}

\textbf{On the Main Components.}
We investigate the effect of the proposed main components, including cross-layer dynamic connection (CDC), dynamic aggregation (DA) and attentive reinforcement (AR), on three groups of datasets in VTAB-1k. All the results are obtained based on the ViT-B/16 backbone supervised pre-trained on ImageNet-21k. 
As shown in Table ~\ref{diff_components}, CDC (see Eq.~\eqref{no_dyn_residual}) clearly boosts the performance of the baseline method VPT, demonstrating the importance of sharing task-relevant information. By additionally applying the DA as depicted in Eq.~\eqref{dyn_residual} and AR module further improve the accuracy, as they enable more flexible attention process and highlight the salient image regions, respectively.

\begin{table}[t]\small  
  \caption{Comparison of various prompt structures as displayed in Fig.~\ref{arch_ill}, by using the ViT-B/16 backbone supervised pre-trained on ImageNet-21k.}
  \vskip 0.1in
    \centering 
    \setlength{\tabcolsep}{3.3pt}
    \begin{tabular}{l c c c c c} 
        \toprule 
        \makecell[l]{Structure} & Natural & Specialized & Structured & Mean\\
        \midrule
        \makecell[l]{(A) VPT} & 78.48 & 82.43 & 54.98 & 71.97\\
        \makecell[l]{(B) ProVP}  & 79.82$_{\textcolor{blue}{\uparrow 1.34}}$ & 81.44$_{\textcolor{red}{\downarrow 0.99}}$ & 44.61$_{\textcolor{red}{\downarrow 10.37}}$ & 68.62\\
        \makecell[l]{(C) EXPRESS} & 79.69$_{\textcolor{blue}{\uparrow 1.21}}$ & 84.03$_{\textcolor{blue}{\uparrow 1.60}}$ & 54.99$_{\textcolor{blue}{\uparrow 0.01}}$ & 72.90\\
        \makecell[l]{(D) Vanilla CDC} & 82.14$_{\textcolor{blue}{\uparrow 3.66}}$ & 85.09$_{\textcolor{blue}{\uparrow 2.66}}$ & 55.31$_{\textcolor{blue}{\uparrow 0.33}}$ & 74.18\\
        \makecell[l]{(E) CDC}  & \textbf{82.57}$_{\textcolor{blue}{\uparrow 4.09}}$ & \textbf{85.30}$_{\textcolor{blue}{\uparrow 2.87}}$& \textbf{58.37}$_{\textcolor{blue}{\uparrow 3.39}}$ & \textbf{75.41}\\
        \bottomrule 
      \end{tabular}
    \label{arch}
  \end{table}
  
\textbf{On Detailed Design in CDC.} In order to evaluate the advantage of the proposed CDC, we remove the DA and AR module, and compare to the following alternative prompt structures shown in Fig.~\ref{arch_ill}: \textbf{(A)} The baseline method VPT-deep \cite{jia2022vpt}; \textbf{(B)} ProVP \cite{progressive}, which inherits the output prompt tokens and further adds learnable prompt tokens \emph{i.e.} $\hat{\bm{P}}^{l} = \bm{P}^{l}+\mathcal{T}^{l-1}(\hat{\bm{P}}^{l-1})$; \textbf{(C)} EXPRESS \cite{express}, which inherits output prompt tokens and adds learnable prompt tokens after LN, QKV projection and MSA layers; \textbf{(D)} Vanilla CDC, which transfers input prompt tokens from all preceding layers to the current layer, \emph{i.e.} $\hat{\bm{P}}^{l} = \bm{P}^{l}+\hat{\bm{P}}^{l-1}$ or equally $\hat{\bm{P}}^{l} = \bm{P}^{l}+\sum_{i=0}^{l-1} \bm{P}^{i}$; \textbf{(E)} CDC, which transfers input prompt tokens solely from previous layer to the current layer, \emph{i.e.} $\hat{\bm{P}}^{l} = \bm{P}^{l}+\bm{P}^{l-1}$. As summarized in Table~\ref{arch}, we have the following observations:

\textbf{(1)} In most cases, prompt structures with an additive design, including (C), (D), (E) outperform the VPT-deep baseline. This improvement is attributed to the additive structure benefits task-relevant sharing and implicitly adjusts the attention weights, as detailed in Sec.~\ref{proof_residual}. 

\textbf{(2)} Structures (B) and (C) exhibit unstable and sub-optimal performance compared to (D) and (E), particularly in tasks within the specialized and structured group (\emph{e.g.} those involving medical and synthetic images). This performance discrepancy is likely due to a significant domain gap between these tasks and the pre-trained ImageNet dataset. Structures (B) and (C), which inherit output prompt tokens, tend to blend input image information during the attention process. This blending may introduce task-irrelevant noise from the input images, leading to sub-optimal results.

\textbf{(3)} Structure (E) consistently outperforms (D), indicating that task-relevant information has a higher similarity between adjacent layers and therefore transferring only the input prompt tokens from the previous layer yields better performance. Based on these observations, we select structure (E) in our approach.

\begin{figure}[t]
  \centering 
  \includegraphics[width=82mm]{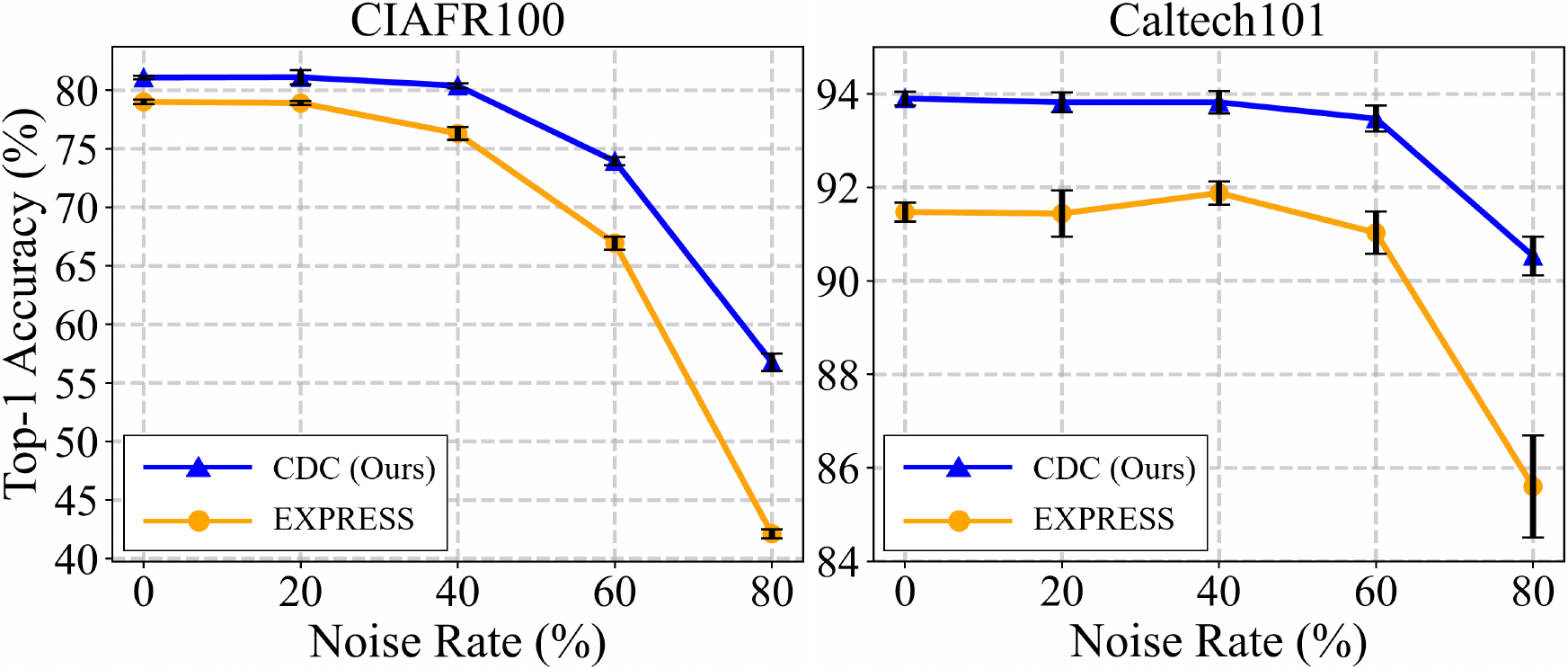}
  \vskip -0.1in
  \caption{Analysis on the robustness of performance against distinct rates of Gaussian noise in the input image, by comparing our CDC method and EXPRESS. } 
  \label{noise}
\end{figure}

\textbf{On the Noise Robustness of CDC.} 
To delve deeper into CDC, we conduct experiments focusing on noise robustness. In this context, we analyze two representative methods for constructing prompt structures based on \emph{input} and \emph{output} prompt tokens, \emph{i.e.} CDC and EXPRESS. We repeat the experiments for three times with distinct random seeds, and
report the mean and standard deviation of the top-1 accuracy on CIFAR100 and Caltech101 datasets. As shown in Fig.~\ref{noise}, CDC exhibits better noise robustness when gradually increasing the proportion of Gaussian noise in the input image. This robustness can be attributed to the unique design of CDC, which not transmits image information through cross-layer connection, effectively minimizing the intrusion of task-irrelevant noise in the images.

\textbf{On Detailed Design in AR.}
We evaluate three settings for the adding position of learnable prompt tokens in AR: \textbf{1)} None: do not add any prompt tokens to the image tokens;
\textbf{2)} All: add to all the image tokens;
\textbf{3)} Top-$k$: add to the top-$k$ most salient image tokens.
As shown in Table~\ref{ar_setting}, adding prompts to all the image tokens (All) outperforms the baseline (None) on 2 out 3 groups in VTAB-1k, showing the necessity of reinforcement. Adding prompts to the most salient token (Top-$k$) for each individual input image reports the best results and steadily outperforms (All) and (None), 
demonstrating the importance of highlighting salient regions based on different input images.

\begin{figure}[t]
  \centering 
  \includegraphics[width=82mm]{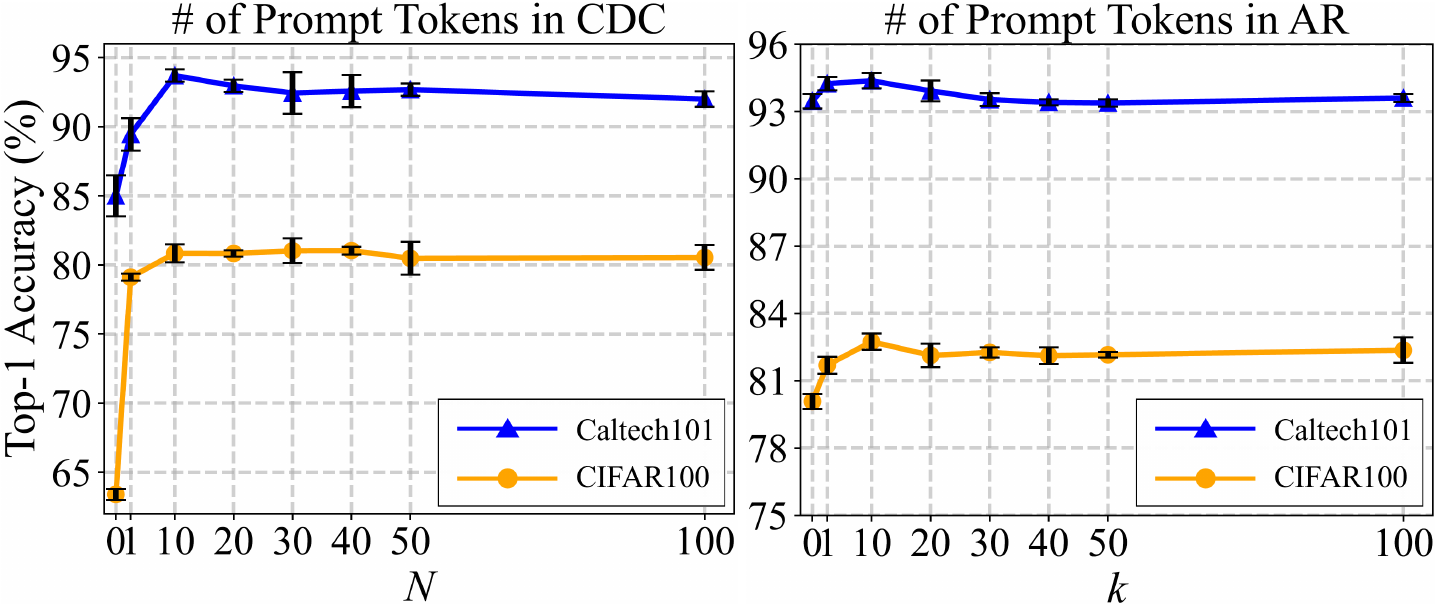}
  \vskip -0.1in
  \caption{Ablation results on the number of prompt tokens ($N$) in CDC and the number of prompt tokens ($k$) in AR.} 
  \label{topk_abla}
  
\end{figure}

\textbf{On Hyper-parameters.} We evaluate the influence of two hyper-parameters on iVPT, \emph{i.e.} the number of prompt tokens in CDC and the number of prompt tokens in AR, denoted by $N$ and $k$, respectively. 
For different settings of $N$, we fix $k$ to 0. 
For different settings of $k$, we fix $N$ to 10.
As shown in Fig,~\ref{topk_abla}, though the optimal choice of $N$ varies for different datasets, iVPT exhibits robustness to hyper-parameter variations. 
In regards of $k$, iVPT constantly reaches better performance compared to $k=0$, \emph{i.e.} without using AR. 
Empirically, setting $N$ and $k$ to 10 is a good choice for initialization, and can be further adjusted by grid searching on the validation set.

Please refer to the \emph{appendix} for more analysis on runtime cost, fine-grained image classification tasks, noise robustness and visualization results. 

\begin{table}[t] \small
  \caption{Comparison of top-1 accuracies (\%) of different settings in AR module.}
  \vskip 0.1in
  \centering%
  \setlength{\tabcolsep}{3.5pt}
  \begin{tabular}{l c c c c} 
      \toprule 
      \makecell[l]{Setting} & Natural & Specialized & Structured & Mean \\
      \midrule
      \makecell[l]{None} & 82.86 & 85.99 & 59.19 & 76.01 \\
      \makecell[l]{All}  & 83.26 & 86.28 & 55.60 & 75.05 \\         
      \makecell[l]{Top-$k$} & \textbf{83.50} & \textbf{87.13} & \textbf{60.81} & \textbf{77.13} \\         
      \bottomrule 
    \end{tabular}
    \label{ar_setting}
\end{table}

\begin{table}[t] \small
    
    \caption{Top-1 accuracies (\%) of iVPT by using distinct backbones supervised pre-trained on ImageNet-21k.}
    \vskip 0.1in
    \centering%
    \setlength{\tabcolsep}{2.5pt}
    \begin{tabular}{l l c c c c} 	
      \toprule 
      \makecell[l]{Backbone} & & Natural & Specialized & Structured & Mean \\
      \midrule
       \multirow{2}{*}{ViT-B/16} & VPT & 78.48 & 82.43 & 54.98 & 71.97 \\
       ~ & \cellcolor{gray!20}iVPT & \cellcolor{gray!20}\textbf{83.50} & \cellcolor{gray!20}\textbf{87.13} & \cellcolor{gray!20}\textbf{60.81} & \cellcolor{gray!20}\textbf{77.13} \\
       \midrule
       \multirow{2}{*}{ViT-L/16} & VPT & 82.48 & 83.87 & 54.13 & 73.49 \\
       ~ & \cellcolor{gray!20}iVPT & \cellcolor{gray!20}\textbf{83.14} & \cellcolor{gray!20}\textbf{84.71} & \cellcolor{gray!20}\textbf{56.14} & \cellcolor{gray!20}\textbf{74.66} \\
       \midrule
       \multirow{2}{*}{ViT-H/14} & VPT & 77.87 & 83.30 & 52.23 & 71.13\\
         & \cellcolor{gray!20}iVPT & \cellcolor{gray!20}\textbf{79.54} & \cellcolor{gray!20}\textbf{84.42} & \cellcolor{gray!20}\textbf{55.62} & \cellcolor{gray!20}\textbf{73.19} \\
        \midrule
        \multirow{2}{*}{Swin-B} & VPT & 76.78 & 84.53 & 53.35 & 71.55\\
         & \cellcolor{gray!20}iVPT & \cellcolor{gray!20}\textbf{82.21} & \cellcolor{gray!20}\textbf{84.87} & \cellcolor{gray!20}\textbf{57.79} & \cellcolor{gray!20}\textbf{74.96} \\

        \bottomrule 
      \end{tabular}
      
      \label{diff_arch}
  \end{table}

\subsection{Generalizability of the Proposed Method}
\label{Generalizability}
We evaluate the generalizability in regards of distinct backbones, pre-training strategies and segmentation task.

\textbf{On the Backbone.}
We adopt the widely used vision transformers including 
ViT-Base (ViT-B/16), ViT-Large (ViT-L/16), ViT-Huge (ViT-H/14) \cite{vit} as well as the hierarchical Swin transformer (Swin-B) \cite{swin}, all of which are supervised pre-trained on ImageNet. As shown in Table~\ref{diff_arch}, iVPT consistently promotes the performance of VPT using ViT backbones with various scales and architectures.

\textbf{On the Pre-training Strategy.}
Besides the supervised pre-training strategy used in Table~\ref{vtab_sup_compare}, 
we investigate the effect of the widely used self-supervised pre-training strategies on iVPT with the ViT-B/16 backbone, including
the contrastive self-supervised method MoCo-v3 \cite{mocov3} and the masked image modeling \cite{beit} method MAE \cite{mae}. In addition to VPT, we additionally compare with GaPT \cite{gate}, which is specifically designed for adapting self-supervised pre-trained models. 
As displayed in Table~\ref{diff_pretrain}, iVPT delivers higher accuracy than VPT and GaPT on VTAB-1k benchmark in regardless of the pre-training strategy, showing its generalizability.

\textbf{On the Semantic Segmentation Task.}
We perform experiments on ADE20K \cite{ade} and 
select SETR \cite{setr} as the segmentation framework, which adds a standard ConvNet head to the ViT-L/16 backbone to perform segmentation. 
Since the segmentation task requires attention to the entire image, we discard the AR module and only add CDC and DA to the ViT backbone. 
As shown in Table~\ref{segment}, iVPT outperforms Linear probing, Bias and VPT with similar parameters using 10 prompt tokens, and further performance gains can be obtained by increasing the number of prompt tokens to 100. 

\begin{table}[t] \small
  \caption{Top-1 accuracies (\%) of iVPT with different pre-training strategies using ViT-B/16.}
  \vskip 0.1in
  \centering%
  \setlength{\tabcolsep}{2.5pt}
  \begin{tabular}{l l c c c c} 	
    \toprule 
    \multicolumn{2}{l}{\makecell[l]{Pre-training 
    strategy}} & Natural & Specialized & Structured & Mean \\
    \midrule
     \multirow{2}{*}{Supervised} & VPT & 78.48 & 82.43 & 54.98 & 71.97 \\
     ~ & \cellcolor{gray!20}iVPT & \cellcolor{gray!20}\textbf{83.50} & \cellcolor{gray!20}\textbf{87.13} & \cellcolor{gray!20}\textbf{60.81} & \cellcolor{gray!20}\textbf{77.13} \\
     \midrule
     \multicolumn{2}{l}{\textcolor{gray}{Self-supervised:}}\\
     \multirow{3}{*}{MoCo-v3} & VPT & 70.27 & 83.04 & 42.38 & 65.23 \\
     ~ & GaPT & 74.84 & 83.38 & 49.10 & 69.11 \\
     ~ & \cellcolor{gray!20}iVPT & \cellcolor{gray!20}\textbf{76.12} & \cellcolor{gray!20}\textbf{84.51} & \cellcolor{gray!20}\textbf{57.88} & \cellcolor{gray!20}\textbf{72.84} \\
     \midrule
     \multirow{3}{*}{MAE} & VPT & 36.02 & 60.61 & 26.57 & 41.07\\
     ~ & GaPT & 47.61 & 76.86 & 36.80 & 53.76 \\
       & \cellcolor{gray!20}iVPT & \cellcolor{gray!20}\textbf{55.05} & \cellcolor{gray!20}\textbf{79.21} & \cellcolor{gray!20}\textbf{50.18} & \cellcolor{gray!20}\textbf{61.48} \\
     
      \bottomrule 
    \end{tabular}
    
    \label{diff_pretrain}
\end{table}

\begin{table}[t] \small
  \caption{Semantic segmentation results on ADE20K \cite{ade} datasets. 'SS/MS' and '$N$' indicate single/multi-scale inference and number of prompt tokens in CDC, respectively.}
  \vskip 0.1in
  \centering%
  \setlength{\tabcolsep}{3.5pt}
  \begin{tabular}{l c c c} 
      \toprule 
      \makecell[l]{Method} & mIoU-SS & mIoU-MS & Parameters (M)\\
      \midrule
      \makecell[l]{Linear probing} & 35.12 & 37.46 & \textbf{13.18} \\
      \makecell[l]{Bias}  & 43.40 & \underline{45.33} & 13.46\\         
      \makecell[l]{VPT} & 42.11 & 44.06 & \underline{13.43} \\
      \rowcolor{gray!20}\makecell[l]{iVPT ($N$=10)~~~} & \underline{43.67} & 45.12 & \underline{13.43} \\
      \rowcolor{gray!20}\makecell[l]{iVPT ($N$=100)} & \textbf{44.77} & \textbf{45.84} & 15.87 \\   
      \bottomrule 
    \end{tabular}
    \label{segment}
    \vspace{-0.5em}
\end{table}

\subsection{Limitations}

Despite of the promising performance, our approach still has several limitations. Firstly, iVPT is exclusively compatible with transformer-based architectures and does not readily extend to convolutional structures. However, the rapidly evolving landscape of foundational models, with a predominant trend favoring transformer-based designs \cite{sam, foundational}, suggests that iVPT holds considerable promise within this context.
Secondly, the AR module is primarily tailored for instance-level tasks. For tasks at the object-level or pixel-level, additional optimizations of the AR module are imperative. This could involve the incorporation of a multi-scale design to facilitate effective attention reinforcement for small-sized objects. Addressing these limitations would further enhance the applicability and versatility of our approach in diverse computer vision tasks.

\section{Conclusion}
In this paper, we propose a novel visual prompt tuning approach, namely
iVPT, for adapting pre-trained vision transformers to downstream tasks. 
iVPT benefits task-relevant information sharing between inter-layer prompt tokens and flexibly modifies the attention process by cross-layer dynamic connection and dynamic aggregation. 
Attentive reinforcement takes advantage of such flexibility and automatically identifies and reinforces the salient image tokens. 
Theoretical analysis and empirical results show the advantage of iVPT among various prompt structures. 
Extensive comparison as well as ablation studies on widely used 
datasets clearly show the effectiveness and generalizability of the proposed approach.

\nocite{langley00}

\bibliography{example_paper}
\bibliographystyle{icml2024}

\newpage
\appendix

\onecolumn

In this appendix, we provide more analysis on iVPT including the runtime cost considerations, fine-grained tasks results, noise robustness results and visualization results in Sec.~\ref{visual}. In addition, we report the results for each dataset in the VTAB benchmark in Sec.~\ref{per_task}, where the average results for each task group are reported in the main body. Finally, we describe more details about the datasets, the setting of hyper-parameters and the pre-trained models used for evaluation in Sec.~\ref{detail}. 

\section{More Analysis on iVPT}
\label{visual}

\begin{table*}[h] \small
  \setlength{\tabcolsep}{3.0pt}
  \caption{Comparison of the runtime cost and accuracy.}
  \vskip 0.1in
  \centering 
      \begin{tabular}{l c c c c c c} 	
      \toprule
      Method & \makecell{Parameters$\downarrow$ \\(M)}  & \makecell{Workload $\downarrow$\\(GFLOPs)} & \makecell{Latency$\downarrow$ \\(Millisecond)}  & \makecell{Training time$\downarrow$ \\(Minute)} & \makecell{Memory$\downarrow$ \\(MB)} & \makecell{Accuracy$\uparrow$ \\(\%)}\\
      \midrule
      \makecell[l]{VPT} & 0.60 & 17.732 & 249.80 & 15.2 & 8,602 & 71.97 \\
      \makecell[l]{DAM-VP} & 2.52 & 33.728 & 763.06 & 19.4 & 11,011 & 73.13 \\
      \makecell[l]{iVPT} & 0.60 & 17.733 & 256.24 & 15.9 & 8,608 & 77.13 \\
      \bottomrule 
      \end{tabular}
      
    \label{runtime} 
  \end{table*}

  \begin{table*}[h] \small
    \setlength{\tabcolsep}{6pt}
    \caption{Comparison of the top-1 accuracies (\%) and learnable parameters (M) by various methods on FGVC benchmark with ViT-B/16 supervised pre-trained on 
    ImageNet-21k. The best and second best results are highlighted in \textbf{bold} and \underline{underline}. `-' indicates the result was not reported. }
    \vskip 0.1in
    \centering 
    \begin{tabular}{l  c  c  c  c  c  c  c} 	
    \toprule
      \makecell[l]{Method} & CUB-200 & NABirds & Oxford Flowers & Stanford Dogs & Stanford Cars & Mean Acc. & Mean Params.\\
      \midrule
      \makecell[l]{Full Fine-tuning} & 87.3 & 82.7 & 98.8 & 89.4 & \underline{84.5} & 88.54 & 85.98\\
      \makecell[l]{Linear Probing} & 85.3 & 75.9 & 97.9 & 86.2 & 51.3 & 79.32 & \textbf{0.18}\\
      \midrule
      \multicolumn{2}{l}{\textcolor{gray}{Prompt-based methods:}}\\
      \makecell[l]{VPT} & \underline{88.5} & 84.2 & 99.0 & 90.2 & 83.6 & 89.11 & 0.85\\
      \makecell[l]{DAM-VP} & 87.5 & 82.1 & \underline{99.2} & \textbf{92.3} & - & - & -\\
      \makecell[l]{EXPRESS} & 88.3 & - & 99.0 & 90.0 & 80.5 & - & -\\
      \makecell[l]{E$^2$VPT} & \textbf{89.1} & \textbf{84.6} & 99.1 & 90.5 & 82.8 & \underline{89.22} & 0.56\\
      \rowcolor{gray!20}\makecell[l]{iVPT ~~~~~~~~~~~~~~~~~~} & \textbf{89.1} & \underline{84.5} & \textbf{99.5} & \underline{90.8} & \textbf{85.6} & \textbf{89.90} & \underline{0.41}\\
  
      \bottomrule
    \end{tabular}
    \label{fgvc} 
  \end{table*}

\textbf{Runtime cost.} 
In Table~\ref{runtime}, we report the mean number of learnable parameters over 19 datasets in VTAB-1k, GFLOPs, inference latency, training time, peak GPU memory footprint and VTAB-1k accuracy for VPT \cite{jia2022vpt}, DAM-VP\cite{dam-vp} and our iVPT. For VPT and iVPT, we adopt the ViT-B/16 backbone and train for 100 epochs on a single NVIDIA 2080Ti GPU. For DAM-VP, we train for 50 epochs as recommended in their original paper. 
Compared to VPT, iVPT introduces cross-layer dynamic connection (CDC), dynamic aggregation (DA) and attentive reinforcement (AR) modules and since the mean learnable parameters within each module remains small, \emph{i.e.} CDC (0.36M), DA (0.02M), AR (0.18M), the overall parameter count (0.60M) remains identical to that of VPT (0.60M). As for the computational cost, iVPT increases negligible computational workload ($\thicksim $ 0.001  GFLOPs), inference latency ($\thicksim$ 10 milliseconds), training time ($\thicksim$ 1 minute) and peak GPU memory footprint ($\thicksim$ 10 MB). However, iVPT delivers remarkably higher accuracy (71.97\% $\rightarrow$ \textbf{77.13}\%), thus reaching a better trade-off between the efficiency and accuracy for deployment. Besides, since the proposed AR module utilizes attention weights to guide learning the prompt tokens via a single forward process without introducing additional query process for prompt selection as in \cite{dam-vp}, iVPT exhibits significant computational advantages compared to DAM-VP~\cite{dam-vp}.

\textbf{Fine-grained task results.} Following \cite{jia2022vpt}, we further evaluate our iVPT on fine-grained visual classification (FGVC) benchmark, which contains five fine-grained image datasets including CUB-200-2011 \cite{cub}, NABirds \cite{nabird}, Oxford Flowers \cite{flowers}, Stanford Dogs \cite{dog} and Stanford Cars \cite{cars}. As shown in Table~\ref{fgvc}, iVPT consistently achieves higher or comparable performance, and reaches the best accuracy on average while requiring the least number of learnable parameters, clearly showing the superiority of our approach on fine-grained tasks. 

\begin{figure*}[t]
  \centering 
  \includegraphics[width=170mm]{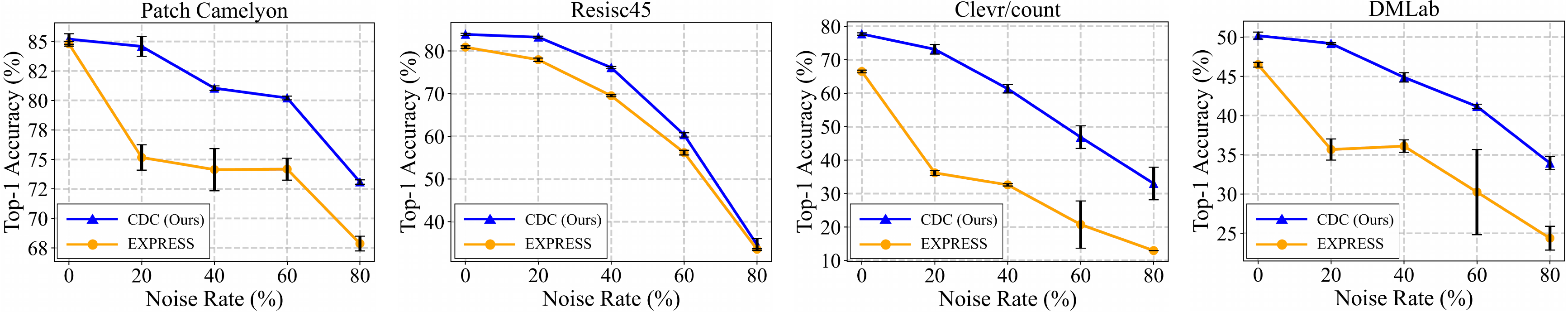}
  \caption{Analysis on the robustness of performance against distinct rates of Gaussian noise in the input image, by comparing our CDC method and EXPRESS. CDC exhibits better noise robustness compared to EXPRESS. }
  \label{supp_noise}
\end{figure*}

\textbf{Noise Robustness results.} In the main body, we conduct the noise robustness analysis on CIAFR100 \cite{cifar} and Caltech101 \cite{cal101} datasets from the natural task group. We further explore the datasets in the specialized and structured task groups and conduct analysis on Patch Camelyon \cite{patch}, Resisc45 \cite{resisc45}, Clevr/count \cite{clevr} and DMLab \cite{dmlab} datasets. As shown in Fig.~\ref{supp_noise}, CDC is more robust to noise in the input image compared to EXPRESS \cite{express}, and this property is more significant on tasks containing medical (Patch Camelyon) and synthetic images (Clevr/count \& DMLab).

\begin{figure*}[t]
  \centering 
  \includegraphics[width=130mm]{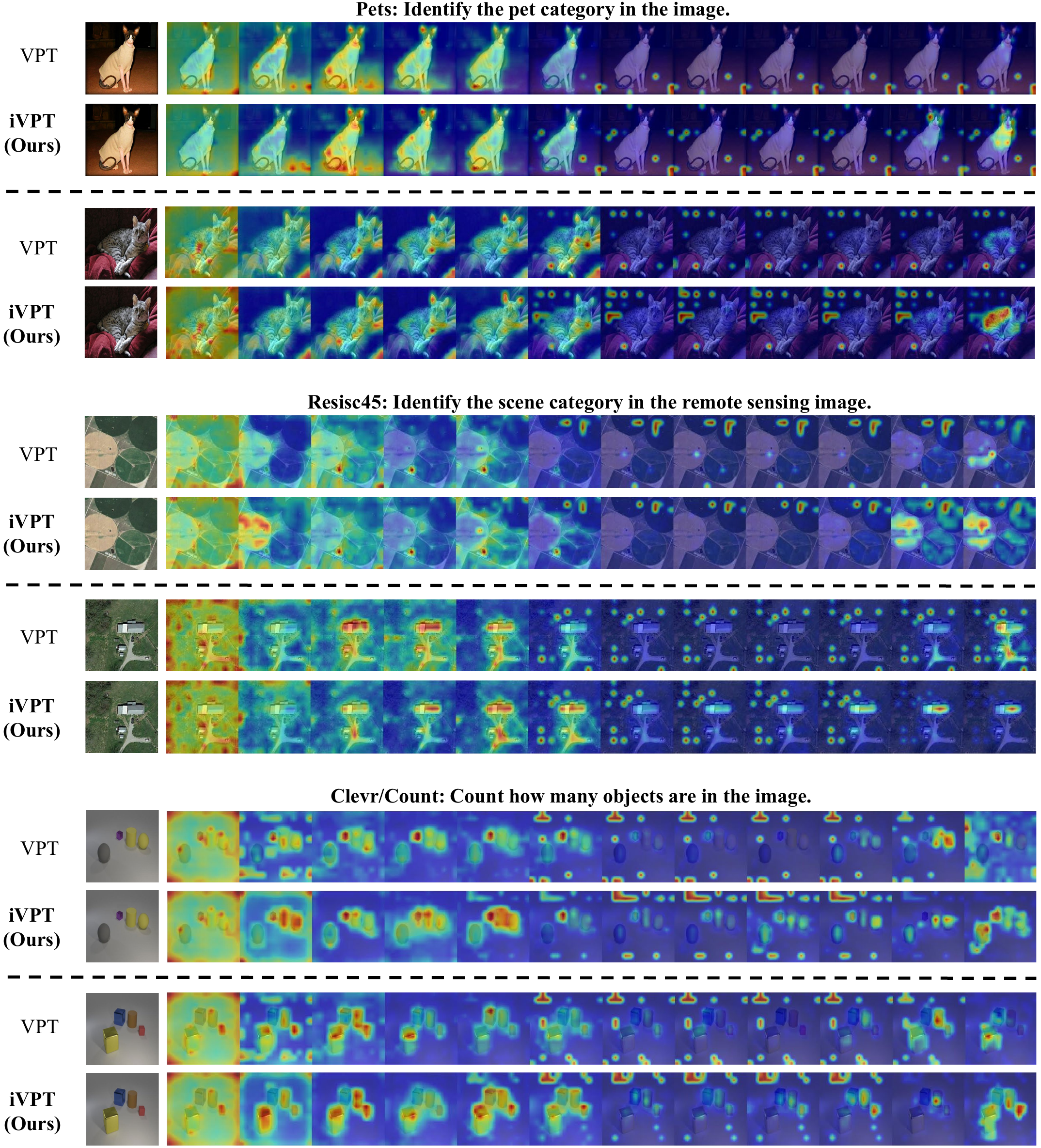}
  \caption{Visualization of attention maps. From left to right, each column shows the original RGB image and the attention maps from the first to the last transformer layers. }
  \label{visual_figure}
\end{figure*}

\textbf{Visualization results.}
We visualize the attention maps from different transform layer by the proposed iVPT approach, compared to VPT \cite{jia2022vpt}. The visualization experiments are conducted on three datasets from different groups in VTAB-1k benchmark, 
including Pets \cite{pets}, Resisc45 \cite{resisc45} and Clevr/Count \cite{clevr}. We adopt the ViT-B/16 backbones supervised pre-trained on ImageNet-21k.  As shown in Fig.~\ref{visual_figure}, for the general image classification tasks on Pets and Resisc45, iVPT generates more concentrated attention on discriminative areas compared with VPT, hence promotes the performance by 3.5\% and 3.7\% on these two datasets, respectively. 
For the object counting task on Clevr/Count,  VPT fails to impose higher attention on some of the foreground objects. In contrast, iVPT clearly focuses on all objects appeared in the scene, thus reaching higher accuracy than VPT, as displayed in Table~1 of the main body.

\newpage

\section{Detailed Results for the VTAB-1k Benchmark.} 
\label{per_task}
In the main body, we report the average results of the VTAB-1k benchmark in some experiments for reasons of space constraints. 
In this section, we report the per-task results in Tables~\ref{diff_comp_per_task}, \ref{diff_arch_per_task},  \ref{diff_set_ar_per_task}, \ref{diff_backbone_per_task} and \ref{diff_pretrain_per_task} corresponding to Tables~\ref{diff_components}, \ref{arch}, \ref{ar_setting}, \ref{diff_arch} and \ref{diff_pretrain} in the main text, respectively.

\begin{table*} \scriptsize
  \centering 
  \setlength{\tabcolsep}{2.6pt}
  \caption{Per task results (\%) of VTAB-1k by various settings in Table~\ref{diff_components} of the main body.} 
  \vskip 0.1in
  \begin{tabular}{c c c c c c c c c c c | c c c c c | c c c c c c c c c | c} 
      \toprule 
      ~ & ~ & ~ & \multicolumn{8}{c}{\textbf{Natural}} & \multicolumn{5}{c}{\textbf{Specialized}} & \multicolumn{9}{c}{\textbf{Structured}} \\
      \cmidrule(lr){4-11}\cmidrule(lr){12-16}\cmidrule(lr){17-25}
      CDC & DA & AR & \rotatebox{90}{CIFAR100} & \rotatebox{90}{Caltech101} & \rotatebox{90}{DTD} & \rotatebox{90}{Flowers102} & 
      \rotatebox{90}{Pets} & \rotatebox{90}{SVHN} & \rotatebox{90}{Sun397} & \rotatebox{90}{Mean Acc.} & 
      \rotatebox{90}{Patch Came.} & \rotatebox{90}{EuroSAT} & \rotatebox{90}{Resisc45} & \rotatebox{90}{Retinopathy} & \rotatebox{90}{Mean Acc.} & 
      \rotatebox{90}{Clevr/count} & \rotatebox{90}{Clevr/dist.} & \rotatebox{90}{DMLab} & \rotatebox{90}{KITTI/dist.} & 
      \rotatebox{90}{dSprites/loc} & \rotatebox{90}{dSprites/ori} & \rotatebox{90}{NORB/azi} & \rotatebox{90}{NORB/ele} & \rotatebox{90}{Mean Acc.} & 
      \rotatebox{90}{Mean Acc.} \\
      \midrule
      \multicolumn{3}{c}{\textcolor{gray}{VPT baseline $\rightarrow$}} & 78.8 & 90.8 & 65.8 & 98.0 & 88.3 & 78.1 & 49.6 & 78.5 & 81.8 & 96.1 & 83.4 & 68.4 & 82.4 & 68.5 & 60.0 & 46.5 & 72.8 & 73.6 & 47.9 & 32.9 & 37.8 & 55.0 & 71.97\\
      \checkmark &  &  & 80.8 & 93.7 & 70.8 & 98.7 & 90.8 & 87.6 & 55.6 & 82.6 & 85.2 & 94.8 & 83.9 & 77.3 & 85.3 & 77.7 & 62.3 & 50.2 & 75.8 & 82.5 & 54.9 & 26.5 & 37.1 & 58.4 & 75.41\\
      \checkmark & \checkmark &  & 80.0 & 93.9 & 71.5 & 98.9 & 91.5 & 89.0 & 55.2 & 82.9 & 86.7 & 95.1 & 84.8 & 77.3 & 86.0 & 73.9 & 62.0 & 50.4 & 78.6 & 87.1 & 56.4 & 28.7 & 36.5 & 59.2 & 76.01\\
      \checkmark & \checkmark & \checkmark & {82.7} & {94.2} & 72.0 & 99.1 & {91.8} & 88.1 & {56.6} & {83.5} & {87.7} & 96.1 & {87.1} & {77.6} & {87.1} & 77.1 & 62.6 & 49.4 & 80.6 & {82.1} & {55.3} & 31.8 & {47.6} & 60.8 & {77.13} \\	
      \bottomrule 
    \end{tabular}
  
  \label{diff_comp_per_task}
\end{table*}

\begin{table*} \scriptsize
  \centering 
  \setlength{\tabcolsep}{2.9pt}
  \caption{Per task results of the top-1 accuracies (\%) by various prompt structures in Table~\ref{arch} of the main body. The best result is highlighted in \textbf{bold}.} 
  \vskip 0.1in
  \begin{tabular}{l c c c c c c c c | c c c c c | c c c c c c c c c | c} 
      \toprule 
      ~ & \multicolumn{8}{c}{\textbf{Natural}} & \multicolumn{5}{c}{\textbf{Specialized}} & \multicolumn{9}{c}{\textbf{Structured}} \\
      \cmidrule(lr){2-9}\cmidrule(lr){10-14}\cmidrule(lr){15-23}
      \makecell[l]{Method} & \rotatebox{90}{CIFAR100} & \rotatebox{90}{Caltech101} & \rotatebox{90}{DTD} & \rotatebox{90}{Flowers102} & 
      \rotatebox{90}{Pets} & \rotatebox{90}{SVHN} & \rotatebox{90}{Sun397} & \rotatebox{90}{Mean Acc.} & 
      \rotatebox{90}{Patch Came.} & \rotatebox{90}{EuroSAT} & \rotatebox{90}{Resisc45} & \rotatebox{90}{Retinopathy} & \rotatebox{90}{Mean Acc.} & 
      \rotatebox{90}{Clevr/count} & \rotatebox{90}{Clevr/dist.} & \rotatebox{90}{DMLab} & \rotatebox{90}{KITTI/dist.} & 
      \rotatebox{90}{dSprites/loc} & \rotatebox{90}{dSprites/ori} & \rotatebox{90}{NORB/azi} & \rotatebox{90}{NORB/ele} & \rotatebox{90}{Mean Acc.} & 
      \rotatebox{90}{Mean Acc.} \\
      \midrule
      \makecell[l]{(A) VPT} & 78.8 & 90.8 & 65.8 & 98.0 & 88.3 & 78.1 & 49.6 & 78.5 & 81.8 & 96.1 & 83.4 & 68.4 & 82.4 & 68.5 & 60.0 & 46.5 & 72.8 & 73.6 & 47.9 & \textbf{32.9} & \textbf{37.8} & 55.0 & 71.97\\
      \makecell[l]{(B) ProVP} & 78.1 & 91.7 & 65.9 & 98.6 & 90.1 & 79.7 & 54.6 & 79.8 & 83.1 & 91.2 & 76.7 & 74.7 & 81.4 & 52.9 & 50.9 & 42.0 & 67.5 & 58.2 & 41.5 & 19.5 & 24.4 & 44.6 & 68.62 \\
      \makecell[l]{(C) EXPRESS} & 78.0 & 89.6 & 68.8 & 98.7 & 88.9 & 81.9 & 51.9 & 79.7 & 84.8 & \textbf{96.2} & 80.9 & 74.2 & 84.0 & 66.5 & 60.4 & 46.5 & \textbf{77.6} & 78.0 & 49.5 & 26.1 & 35.3 & 55.0 & 72.90 \\
      \makecell[l]{(D) Vanilla CDC} & \textbf{82.1} & 93.4 & 68.6 & \textbf{99.0} & \textbf{90.9} & 85.2 & \textbf{55.8} & 82.1 & \textbf{85.4} & 94.2 & \textbf{84.2} & 76.5 & 85.1 & 69.7 & 58.0 & 47.5 & 76.2 & 81.5 & 53.4 & 21.2 & 35.0 & 55.3 & 74.18\\
      \makecell[l]{(E) CDC} & 80.8 & \textbf{93.7} & \textbf{70.8} & 98.7 & 90.8 & \textbf{87.6} & 55.6 & \textbf{82.6} & 85.2 & 94.8 & 83.9 & \textbf{77.3} & \textbf{85.3} & \textbf{77.7} & \textbf{62.3} & \textbf{50.2} & 75.8 & \textbf{82.5} & \textbf{54.9} & 26.5 & 37.1 & \textbf{58.4} & \textbf{75.41}\\
      \bottomrule 
    \end{tabular}
  
  \label{diff_arch_per_task}
\end{table*}


\begin{table*} \scriptsize
  \centering 
  \setlength{\tabcolsep}{2.9pt}
  \caption{Per task results (\%) of VTAB-1k by various settings in AR module in Table~\ref{ar_setting} of the main body.} 
  \vskip 0.1in
  \begin{tabular}{l c c c c c c c c | c c c c c | c c c c c c c c c | c} 
      \toprule 
      ~ & \multicolumn{8}{c}{\textbf{Natural}} & \multicolumn{5}{c}{\textbf{Specialized}} & \multicolumn{9}{c}{\textbf{Structured}} \\
      \cmidrule(lr){2-9}\cmidrule(lr){10-14}\cmidrule(lr){15-23}
      \makecell[l]{Setting} & \rotatebox{90}{CIFAR100} & \rotatebox{90}{Caltech101} & \rotatebox{90}{DTD} & \rotatebox{90}{Flowers102} & 
      \rotatebox{90}{Pets} & \rotatebox{90}{SVHN} & \rotatebox{90}{Sun397} & \rotatebox{90}{Mean Acc.} & 
      \rotatebox{90}{Patch Came.} & \rotatebox{90}{EuroSAT} & \rotatebox{90}{Resisc45} & \rotatebox{90}{Retinopathy} & \rotatebox{90}{Mean Acc.} & 
      \rotatebox{90}{Clevr/count} & \rotatebox{90}{Clevr/dist.} & \rotatebox{90}{DMLab} & \rotatebox{90}{KITTI/dist.} & 
      \rotatebox{90}{dSprites/loc} & \rotatebox{90}{dSprites/ori} & \rotatebox{90}{NORB/azi} & \rotatebox{90}{NORB/ele} & \rotatebox{90}{Mean Acc.} & 
      \rotatebox{90}{Mean Acc.} \\
      \midrule
      \makecell[l]{None} & 80.0 & 93.9 & 71.5 & 98.9 & 91.5 & 89.0 & 55.2 & 82.9 & 86.7 & 95.1 & 84.8 & 77.3 & 86.0 & 73.9 & 62.0 & 50.4 & 78.6 & 87.1 & 56.4 & 28.7 & 36.5 & 59.2 & 76.01\\
      \makecell[l]{All} & 82.4 & 93.7 & 70.5 & 99.1 & 91.7 & 88.8 & 56.6 & 83.3 & 87.3 & 95.1 & 85.2 & 77.5 & 86.3 & 67.9 & 60.8 & 50.7 & 70.6 & 77.9 & 53.5 & 27.2 & 36.1 & 55.6 & 75.05\\
      \makecell[l]{Top-$k$} & {82.7} & {94.2} & 72.0 & 99.1 & {91.8} & 88.1 & {56.6} & {83.5} & {87.7} & 96.1 & {87.1} & {77.6} & {87.1} & 77.1 & 62.6 & 49.4 & 80.6 & {82.1} & {55.3} & 31.8 & {47.6} & 60.8 & {77.13} \\	

      \bottomrule 
    \end{tabular}
  
  \label{diff_set_ar_per_task}
\end{table*}

\begin{table*} \scriptsize
  \centering 
  \setlength{\tabcolsep}{2.9pt}
  \caption{Per task results (\%) of VTAB-1k by distinct backbones supervised pre-trained on ImageNet-21k in Table~\ref{diff_arch} of the main body. The best result is highlighted in \textbf{bold}. `-' indicates the result was not reported.} 
  \vskip 0.1in
  \begin{tabular}{c l c c c c c c c c | c c c c c | c c c c c c c c c | c} 
      \toprule 
      ~ &~ & \multicolumn{8}{c}{\textbf{Natural}} & \multicolumn{5}{c}{\textbf{Specialized}} & \multicolumn{9}{c}{\textbf{Structured}} \\
      \cmidrule(lr){3-10}\cmidrule(lr){11-15}\cmidrule(lr){16-24}
      \makecell[l]{Backbone} & ~ & \rotatebox{90}{CIFAR100} & \rotatebox{90}{Caltech101} & \rotatebox{90}{DTD} & \rotatebox{90}{Flowers102} & 
      \rotatebox{90}{Pets} & \rotatebox{90}{SVHN} & \rotatebox{90}{Sun397} & \rotatebox{90}{Mean Acc.} & 
      \rotatebox{90}{Patch Came.} & \rotatebox{90}{EuroSAT} & \rotatebox{90}{Resisc45} & \rotatebox{90}{Retinopathy} & \rotatebox{90}{Mean Acc.} & 
      \rotatebox{90}{Clevr/count} & \rotatebox{90}{Clevr/dist.} & \rotatebox{90}{DMLab} & \rotatebox{90}{KITTI/dist.} & 
      \rotatebox{90}{dSprites/loc} & \rotatebox{90}{dSprites/ori} & \rotatebox{90}{NORB/azi} & \rotatebox{90}{NORB/ele} & \rotatebox{90}{Mean Acc.} & 
      \rotatebox{90}{Mean Acc.} \\
      \midrule
      \multirow{2}{*}{ViT-B/16} & VPT & 78.8 & 90.8 & 65.8 & 98.0 & 88.3 & 78.1 & 49.6 & 78.5 & 81.8 & \textbf{96.1} & 83.4 & 68.4 & 82.4 & 68.5 & 60.0 & 46.5 & 72.8 & 73.6 & 47.9 & \textbf{32.9} & 37.8 & 55.0 & 71.97\\
       & iVPT & \textbf{82.7} & \textbf{94.2} & \textbf{72.0} & \textbf{99.1} & \textbf{91.8} & \textbf{88.1} & \textbf{56.6} & \textbf{83.5} & \textbf{87.7} & \textbf{96.1} & \textbf{87.1} & \textbf{77.6} & \textbf{87.1} & \textbf{77.1} & \textbf{62.6} & \textbf{49.4} & \textbf{80.6} & \textbf{82.1} & \textbf{55.3} & {31.8} & \textbf{47.6} & \textbf{60.8} & \textbf{77.13} \\
      \midrule
      \multirow{2}{*}{ViT-L/16} & VPT & 84.1 & 88.9 & 70.8 & 98.8 & 90.0 & \textbf{89.0} & 55.9 & 82.5 & 82.5 & \textbf{96.6} & 82.6 & 73.9 & 83.9 & 63.7 & 60.7 & \textbf{46.1} & 75.7 & 83.7 & 47.4 & \textbf{18.9} & \textbf{36.9} & 54.1 & 73.49\\
       & iVPT & \textbf{86.8} & \textbf{91.8} & \textbf{71.1} & \textbf{99.0} & \textbf{90.7} & {85.8} & \textbf{56.9} & \textbf{83.1} & \textbf{82.9} & {95.2} & \textbf{85.0} & \textbf{75.8} & \textbf{84.7} & \textbf{73.3} & \textbf{61.0} & {46.0} & \textbf{79.8} & \textbf{86.7} & \textbf{52.2} & {18.5} & {31.7} & \textbf{56.1} & \textbf{74.66}\\
      \midrule
      \multirow{2}{*}{ViT-H/14} & VPT & 76.9 & 87.2 & 66.8 & 97.5 & 84.8 & \textbf{85.5} & 46.5 & 77.9 & 81.6 & \textbf{96.3} & 82.5 & 72.8 & 83.3 & 50.4 & 61.2 & 43.9 & \textbf{76.6} & 79.5 & 50.1 & 24.7 & 31.5 & 52.2 & 71.13\\
       & iVPT & \textbf{77.3} & \textbf{91.8} & \textbf{68.6} & \textbf{97.8} & \textbf{88.6} & {84.9} & \textbf{47.7} & \textbf{79.5} & \textbf{83.7} & {95.3} & \textbf{83.4} & \textbf{75.3} & \textbf{84.4} & \textbf{63.7} & \textbf{61.3} & \textbf{47.2} & {75.1} & \textbf{79.6} & \textbf{51.5} & \textbf{33.0} & \textbf{33.5} & \textbf{55.6} & \textbf{73.19}\\
      \midrule
      \multirow{2}{*}{Swin-B} & VPT & - & - & - & - & - & - & - & 76.8 & - & - & - & - & 84.5 & - & - & - & - & - & - & - & - & 53.4 & 71.55\\
       & iVPT & 80.8 & 91.3 & 76.8 & 99.6 & 89.9 & 86.3 & 50.8 & \textbf{82.2} & 84.3 & 94.4 & 84.8 & 76.0 & \textbf{84.9} & 76.4 & 62.1 & 73.7 & 68.8 & 75.7 & 54.1 & 22.3 & 29.3 & \textbf{57.8} & \textbf{74.96} \\
      \bottomrule 
    \end{tabular}
  
  \label{diff_backbone_per_task}
\end{table*}

\begin{table*} \scriptsize
  \centering 
  \setlength{\tabcolsep}{2.9pt}
  \caption{Per task results (\%) of VTAB-1k by using different pre-training strategies in Table~\ref{diff_pretrain} of the main body. The best result is highlighted in \textbf{bold}. `-' indicates the result was not reported.} 
  \vskip 0.1in
  \begin{tabular}{l l c c c c c c c c | c c c c c | c c c c c c c c c | c} 
      \toprule 
      ~ &~ & \multicolumn{8}{c}{\textbf{Natural}} & \multicolumn{5}{c}{\textbf{Specialized}} & \multicolumn{9}{c}{\textbf{Structured}} \\
      \cmidrule(lr){3-10}\cmidrule(lr){11-15}\cmidrule(lr){16-24}
      \makecell[l]{Backbone} & ~ & \rotatebox{90}{CIFAR100} & \rotatebox{90}{Caltech101} & \rotatebox{90}{DTD} & \rotatebox{90}{Flowers102} & 
      \rotatebox{90}{Pets} & \rotatebox{90}{SVHN} & \rotatebox{90}{Sun397} & \rotatebox{90}{Mean Acc.} & 
      \rotatebox{90}{Patch Came.} & \rotatebox{90}{EuroSAT} & \rotatebox{90}{Resisc45} & \rotatebox{90}{Retinopathy} & \rotatebox{90}{Mean Acc.} & 
      \rotatebox{90}{Clevr/count} & \rotatebox{90}{Clevr/dist.} & \rotatebox{90}{DMLab} & \rotatebox{90}{KITTI/dist.} & 
      \rotatebox{90}{dSprites/loc} & \rotatebox{90}{dSprites/ori} & \rotatebox{90}{NORB/azi} & \rotatebox{90}{NORB/ele} & \rotatebox{90}{Mean Acc.} & 
      \rotatebox{90}{Mean Acc.} \\
      \midrule
      \multirow{2}{*}{Supervised} & VPT & 78.8 & 90.8 & 65.8 & 98.0 & 88.3 & 78.1 & 49.6 & 78.5 & 81.8 & \textbf{96.1} & 83.4 & 68.4 & 82.4 & 68.5 & 60.0 & 46.5 & 72.8 & 73.6 & 47.9 & \textbf{32.9} & 37.8 & 55.0 & 71.97\\
       & iVPT & \textbf{82.7} & \textbf{94.2} & \textbf{72.0} & \textbf{99.1} & \textbf{91.8} & \textbf{88.1} & \textbf{56.6} & \textbf{83.5} & \textbf{87.7} & \textbf{96.1} & \textbf{87.1} & \textbf{77.6} & \textbf{87.1} & \textbf{77.1} & \textbf{62.6} & \textbf{49.4} & \textbf{80.6} & \textbf{82.1} & \textbf{55.3} & {31.8} & \textbf{47.6} & \textbf{60.8} & \textbf{77.13} \\
      \midrule
      \multicolumn{2}{l}{\textcolor{gray}{Self-supervised:}}\\
      \multirow{3}{*}{MoCo-v3} & VPT & - & - & - & - & - & - & - & 70.3 & - & - & - & - & 83.0 & - & - & - & - & - & - & - & - & 42.4 & 65.23\\
       & GaPT & - & - & - & - & - & - & - & 74.8 & - & - & - & - & 83.4 & - & - & - & - & - & - & - & - & 49.1 & 69.11\\
       & iVPT & 73.1 & 92.2 & 66.7 & 87.8 & 88.8 & 84.9 & 39.5 & \textbf{76.1} & 87.6 & 95.3 & 78.3 & 76.9 & \textbf{84.5} & 72.3 & 63.1 & 47.9 & 81.7 & 82.5 & 48.4 & 23.4 & 43.8 & \textbf{57.9} & \textbf{72.84}\\
      \midrule

      \multirow{3}{*}{MAE} & VPT & - & - & - & - & - & - & - & 36.0 & - & - & - & - & 60.6 & - & - & - & - & - & - & - & - & 26.6 & 41.07\\
       & GaPT & - & - & - & - & - & - & - & 47.6 & - & - & - & - & 76.9 & - & - & - & - & - & - & - & - & 36.8 & 53.76\\
       & iVPT & 34.0 & 82.9 & 62.1 & 78.9 & 83.8 & 19.8 & 23.8 & \textbf{55.1} & 75.5 & 91.8 & 75.1 & 74.4 & \textbf{79.2} & 76.3 & 60.5 & 46.5 & 79.9 & 82.2 & 15.9 & 12.7 & 27.5 & \textbf{50.2} & \textbf{61.48}\\
      \bottomrule 
    \end{tabular}
  
  \label{diff_pretrain_per_task}
\end{table*}

\section{More Implementation Details}
\label{detail}
We adopt almost the same settings of dataset as depicted in VPT \cite{jia2022vpt}. The specific statistics of the 19 datasets in VTAB-1k, 5 datasets in FGVC and 1 dataset in semantic segmentation benchmark are summarized in Table~\ref{datasets}, including the number of classes, the number of images used for training, validation and testing. We also report the hyper-parameters, \emph{i.e.} the numbers of prompt tokens in CDC and AR, corresponding to Table 1 of the main body in Table~\ref{hyper-params} and Table~\ref{fgvc} of the supplementary material in Table~\ref{hyper_fgvc}. Additionally, we display the pre-trained strategy, the pre-training dataset, the number of parameters, the dimension $D$ of tokens in the last transform layer, and the publicly released download link of the pre-trained checkpoints in Table~\ref{pretrained-model}. 

\begin{table}[t] \small
  \setlength{\tabcolsep}{5pt}
  \caption{Specifications of the hyper-parameters used in Table~\ref{fgvc} of the supplementary material. 
  $N$: the number of prompt tokens in CDC. $k$: the number of the prompt tokens in AR.}
  \vskip 0.1in
  \centering 
  \begin{tabular}{l  c  c  c  c  c} 	
  \toprule
    \makecell[l]{Hyper- \\ parameters} & {CUB} & {NABirds} & {Flowers} & {Dogs} & {Cars} \\
    \midrule
    \makecell[l]{$N$} & 10 & 50 & 5 & 10 & 10 \\
    \makecell[l]{$k$} & 1 & 10 & 10 & 10 & 10 \\
    \bottomrule
  \end{tabular}
  
  \label{hyper_fgvc} 
\end{table}

\begin{table*}[t]\small 
\centering 
\setlength{\tabcolsep}{4.5pt}
\caption{Specifications of the hyper-parameters used in Table 1 of the main body. 
$N$: the number of prompt tokens in CDC. $k$: the number of the prompt tokens in AR.} 
\vskip 0.1in
\begin{tabular}{l c c c c c c c | c c c c | c c c c c c c c } 
    \toprule 
    ~ & \multicolumn{7}{c}{\textbf{Natural}} & \multicolumn{4}{c}{\textbf{Specialized}} & \multicolumn{8}{c}{\textbf{Structured}} \\
    \cmidrule(lr){2-8}\cmidrule(lr){9-12}\cmidrule(lr){13-20}
    \makecell[l]{Hyper- \\ parameters} & \rotatebox{90}{CIFAR100} & \rotatebox{90}{Caltech101} & \rotatebox{90}{DTD} & \rotatebox{90}{Flowers102} & 
    \rotatebox{90}{Pets} & \rotatebox{90}{SVHN} & \rotatebox{90}{Sun397} &  
    \rotatebox{90}{Patch Came.} & \rotatebox{90}{EuroSAT} & \rotatebox{90}{Resisc45} & \rotatebox{90}{Retinopathy} &  
    \rotatebox{90}{Clevr/count} & \rotatebox{90}{Clevr/dist.} & \rotatebox{90}{DMLab} & \rotatebox{90}{KITTI/dist.} & 
    \rotatebox{90}{dSprites/loc} & \rotatebox{90}{dSprites/ori} & \rotatebox{90}{NORB/azi} & \rotatebox{90}{NORB/ele}\\
    \midrule
    \makecell[l]{$N$} & 10 & 10 & 10 & 10 & 10 & 50 & 5 & 30 & 50 & 10 & 30 & 100 & 50 & 10 & 50 & 10 & 50 & 50 & 200  \\
    \makecell[l]{$k$} & 10 & 5 & 10 & 50 & 20 & 20 & 1 & 30 & 5 & 50 & 30 & 50 & 100 & 10 & 10 & 1 & 1 & 10 & 1  \\
    
    \bottomrule 
  \end{tabular}
\label{hyper-params}
\end{table*}

\begin{table*}[t] \small
  \setlength{\tabcolsep}{6pt}
  \caption{Specifications of the datasets for evaluation. The statistics below were reported in \cite{jia2022vpt}. }
  \vskip 0.1in
	\centering 
	\begin{tabular}{l  l  l  l  l  l } 	
	\toprule
    \makecell[l]{Dataset} &  Desctiption & \# Classes & Train & Val & Test\\
    \midrule
    \multicolumn{2}{l}{\textcolor{gray}{Visual task adaptation benchmark (VTAB-1k):}}\\
    \makecell[l]{CIFAR100 \cite{cifar}} & \multirow{7}{*}{Natural} & 100 & \multirow{7}{*}{1000} & \multirow{7}{*}{200} & 10,000\\
    \makecell[l]{Caltech101 \cite{cal101}} &  & 102 &  &  & 6,084\\
    \makecell[l]{DTD \cite{DTD}} &  & 47 &  &  & 1,880\\
    \makecell[l]{Flowers102 \cite{flowers}} &  & 102 &  &  & 6,149\\
    \makecell[l]{Pets \cite{pets}} &  & 37 &  &  & 3,669\\
    \makecell[l]{SVHN \cite{svhn}} &  & 10 &  &  & 26,032\\
    \makecell[l]{Sun397 \cite{sun397}} &  & 397 &  &  & 21,750\\
    \midrule
    \makecell[l]{Patch Camelyon \cite{patch}} & \multirow{4}{*}{Specialized} & 2 & \multirow{4}{*}{1000} & \multirow{4}{*}{200} & 32,768\\
    \makecell[l]{EuroSAT \cite{eurosat}} &  & 10 &  &  & 5,400\\
    \makecell[l]{Resisc45 \cite{resisc45}} &  & 45 &  &  & 6,300\\
    \makecell[l]{Retinopathy \cite{retino}} &  & 5 &  &  & 42,670\\
    \midrule
    \makecell[l]{Clevr/count \cite{clevr}} & \multirow{8}{*}{Structured} & 8 & \multirow{8}{*}{1000} & \multirow{8}{*}{200} & 15,000\\
    \makecell[l]{Clevr/distance \cite{clevr}} &  & 6 &  &  & 15,000\\
    \makecell[l]{DMLab \cite{dmlab}} &  & 6 &  &  & 22,735\\
    \makecell[l]{KITTI/dstance \cite{kitti}} &  & 4 &  &  & 711\\
    \makecell[l]{dSprites/location \cite{dsprites17}} &  & 16 &  &  & 73,728\\
    \makecell[l]{dSprites/orientation \cite{dsprites17}} &  & 16 &  &  & 73,728\\
    \makecell[l]{SmallNORB/azimuth \cite{norb}} &  & 18 &  &  & 12,150\\
    \makecell[l]{SmallNORB/elevation \cite{norb}} &  & 9 &  &  & 12,150\\
    \midrule
    \multicolumn{2}{l}{\textcolor{gray}{Fine-grained visual classification benchmark (FGVC):}}\\
    \makecell[l]{CUB-200-2011 \cite{cub}} & Bird species recognition & 200 & 5,394 & 600 & 5,794\\
    \makecell[l]{NABirds \cite{nabird}} & Bird species recognition & 555 & 21,536 & 2,393 & 24,633\\
    \makecell[l]{Oxford Flowers \cite{flowers}} & Flower species recognition & 102 & 1,020 & 1,020 & 6,149\\
    \makecell[l]{Stanford Dogs \cite{dog}} & Dog species recognition & 120 & 10,800 & 1,200 & 8,580\\
    \makecell[l]{Stanford Cars \cite{cars}} & Car species recognition & 196 & 7,329 & 815 & 8,041\\
    \midrule
    \multicolumn{2}{l}{\textcolor{gray}{Semantic segmentation benchmark:}}\\
    ADE20k \cite{ade} & Scene and object segmentation & 150 & 20,210 & 2000 & 2000 \\
    \bottomrule
	\end{tabular}
  
  \label{datasets} 
\end{table*}

\begin{table*}[t] \small
  \setlength{\tabcolsep}{6pt}
  \caption{Specifications of different pre-trained backbones used in the paper. `$^*$' indicates the download link publicly released by published works.}
  \vskip 0.1in
	\centering 
	\begin{tabular}{l  c  c  c  c  c } 	
	\toprule
    \makecell[l]{Backbone} & Pre-trained strategy & Pre-trained dataset & Params. (M) & Token dim. $D$ & Model\\
    \midrule
    \makecell[l]{ViT-B/16} & \multirow{3}{*}{Supervised} & \multirow{3}{*}{ImageNet-21k} & 85 & 768 & \href{https://storage.googleapis.com/vit_models/imagenet21k/ViT-B_16.npz}{checkpoint}$^*$\\
    \makecell[l]{ViT-L/16} &  &  & 307 & 1024 & \href{https://storage.googleapis.com/vit_models/imagenet21k/ViT-L_16.npz}{checkpoint}$^*$\\
    \makecell[l]{ViT-H/14} &  &  & 630 & 1280 & \href{https://storage.googleapis.com/vit_models/imagenet21k/ViT-H_14.npz}{checkpoint}$^*$\\
    \midrule
    \makecell[l]{ViT-B/16} & MoCo-v3 & \multirow{2}{*}{ImageNet-1k} & \multirow{2}{*}{85} & \multirow{2}{*}{768} & \href{https://dl.fbaipublicfiles.com/moco-v3/vit-b-300ep/linear-vit-b-300ep.pth.tar}{checkpoint}$^*$\\
    \makecell[l]{ViT-B/16} & MAE &  &  &  & \href{https://dl.fbaipublicfiles.com/mae/pretrain/mae_pretrain_vit_base.pth}{checkpoint}$^*$\\
    \midrule
    \makecell[l]{Swin-B} & Supervised & ImageNet-21k & 88 & 1024 & \href{https://github.com/SwinTransformer/storage/releases/download/v1.0.0/swin_base_patch4_window7_224_22k.pth}{checkpoint}$^*$\\
    \bottomrule
	\end{tabular}
  
  \label{pretrained-model} 
\end{table*}

\end{document}